\pdfoutput=1

\documentclass[11pt]{article}

\usepackage[final]{acl}
\usepackage{times}
\usepackage{latexsym}
\usepackage{booktabs}
\usepackage{multirow}
\usepackage{xcolor}
\usepackage{amsmath}
\usepackage{amssymb}
\usepackage{CJK}
\usepackage{array}   

\usepackage[T1]{fontenc}
\usepackage{graphicx}
\usepackage{enumitem}
\usepackage{tikz}

\usepackage{makecell}
\usepackage{hyperref}
\usepackage{url}
\usepackage{booktabs}          
\usepackage{siunitx}           
\usepackage{caption}           
\usepackage{adjustbox}         
\usepackage[skip=0.5ex]{subcaption}
\usepackage{booktabs}   
\usepackage{siunitx}    
\usepackage{multirow}   
\usepackage[table]{xcolor}
\usepackage[ruled,vlined]{algorithm2e} 

\usepackage{algorithmic}
\usepackage{amsmath}
\usepackage[table]{xcolor}
\usepackage{booktabs}
\usepackage{multirow}
\usepackage{caption}
\usepackage{enumitem}

\newcommand*\circled[1]{\tikz[baseline=(char.base)]
  \node[shape=circle,draw,line width=1pt,inner sep=0.8pt] (char) {#1};}

\usepackage[utf8]{inputenc}

\usepackage{microtype}

\usepackage{inconsolata}

\usepackage{graphicx}
\usepackage{multirow}
\usepackage{booktabs}
\usepackage{graphicx}
\usepackage{subcaption}
\title{Unlocking Multilingual Reasoning Capabilities of LLMs and LVLMs via Representation Engineering}

\author{Qiming Li$^{1,*}$, Xiaocheng Feng$^{1,2}$, Yixuan Ma$^1$\thanks{Equal contribution.}, Ruihan Chen$^1$, Zihe Tong$^1$, Zekai Ye$^1$ \\
\textbf{Xiachong Feng$^3$, Libo Qin$^4$, Haoyu Ren$^5$, Kun Chen$^5$, Yunfei Lu$^5$, Dandan Tu$^5$, Bing Qin$^{1,2}$}\\
  $^{1}$Harbin Institute of Technology\quad $^2$Peng Cheng Laboratory \quad $^3$The University of Hong Kong\\
  $^4$Harbin Institute of Technology, Shenzhen\quad$^5$Huawei Technologies Co., Ltd
  \\
  \texttt{qmli@ir.hit.edu.cn}
  \\
}

\begin{document}
\maketitle
\begin{abstract}
Large Language Models (LLMs) and Large Vision-Language Models (LVLMs) demonstrate strong reasoning capabilities, yet their performance in English significantly outperforms that in low-resource languages, raising fairness concerns in multilingual applications. Existing approaches either rely on costly multilingual training or employ prompting with external translation tools, both of which are resource-intensive and sensitive to translation quality. 
To address these limitations, we propose a training-free inference-time method to enhance \textbf{M}ultilingual \textbf{R}easoning capabilities via \textbf{R}epresentation \textbf{E}ngineering (\textbf{MRRE}) without using any additional training data or tools. \textbf{MRRE} sequentially injects two precomputed vectors at specific layers during inference processing: cross-lingual reasoning enhancement vectors, which steer non-English reasoning representations toward English space to unlock multilingual reasoning, and target-language output anchoring vectors, which restore the distribution of the target language to preserve input–output language consistency. Comprehensive experiments across six advanced LLMs and LVLMs on four reasoning benchmarks demonstrate MRRE consistently enhances non-English reasoning by an average gain of 5.48\% and up to 7.54\% in low-resource languages (e.g., Thai and Swahili), while improving input-output language consistency by 3.78\%.
\end{abstract}

\section{Introduction}

With the rapid development of Large Language Models (LLMs) and Large Vision-Language Models (LVLMs), foundational models such as Llama-3.1-8B-Instruct \cite{grattafiori2024llama} and Qwen2.5-VL-7B-Instruct \cite{qwen2.5-VL} have demonstrated impressive capabilities in complex reasoning.
However, high-resource languages such as English exhibit substantially stronger reasoning capabilities compared to low-resource languages, raising concerns about fairness in multilingual applications under global deployment.
To address the above issues, prior works primarily focus on two directions to enhance non-English reasoning capabilities:
(1) Data-driven training methods, which align multilingual embeddings \cite{arora2024towards} or construct multilingual reasoning datasets for instruction tuning \cite{fan2025slam}, but inevitably depend on expensive data and incur considerable computational costs. (2) Prompting-based methods, which rely on external translation tools or models \cite{khandelwal2024moral,liu2024translation}, but are highly sensitive to translation quality and prompt design, accompanied by high latency.
Moreover, existing methods are rarely effective across both LLMs and LVLMs.
Therefore, establishing a unified and efficient paradigm for enhancing multilingual reasoning across both LLMs and LVLMs remains further exploration.
\begin{figure}
  \includegraphics[width=1.0\linewidth]{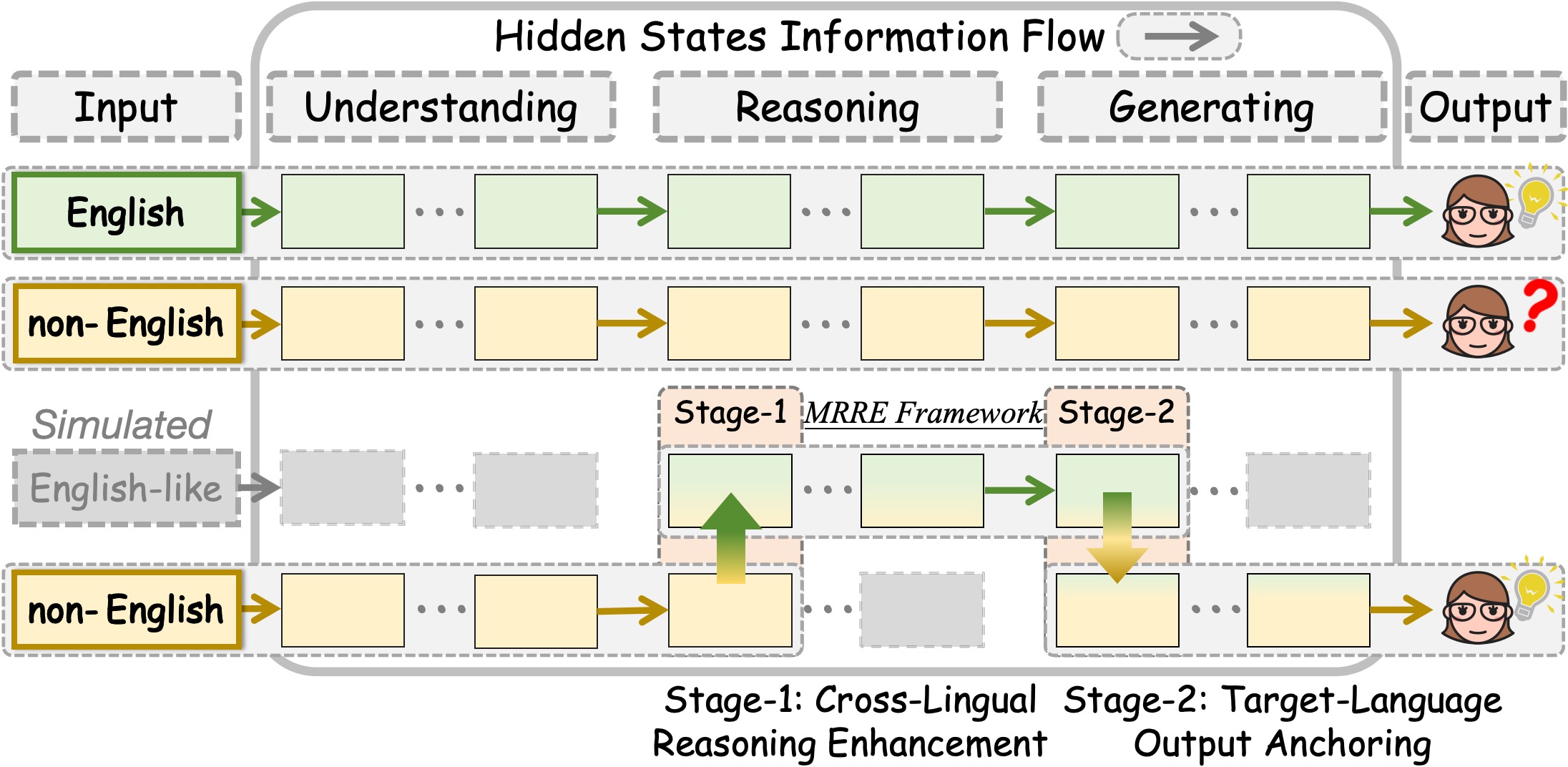}
  \caption{MRRE adopts a two-stage intervention strategy to unlock multilingual reasoning capabilities.}
  \label{fig:pre}
\end{figure}

\begin{figure*}
  \includegraphics[width=0.96\linewidth]{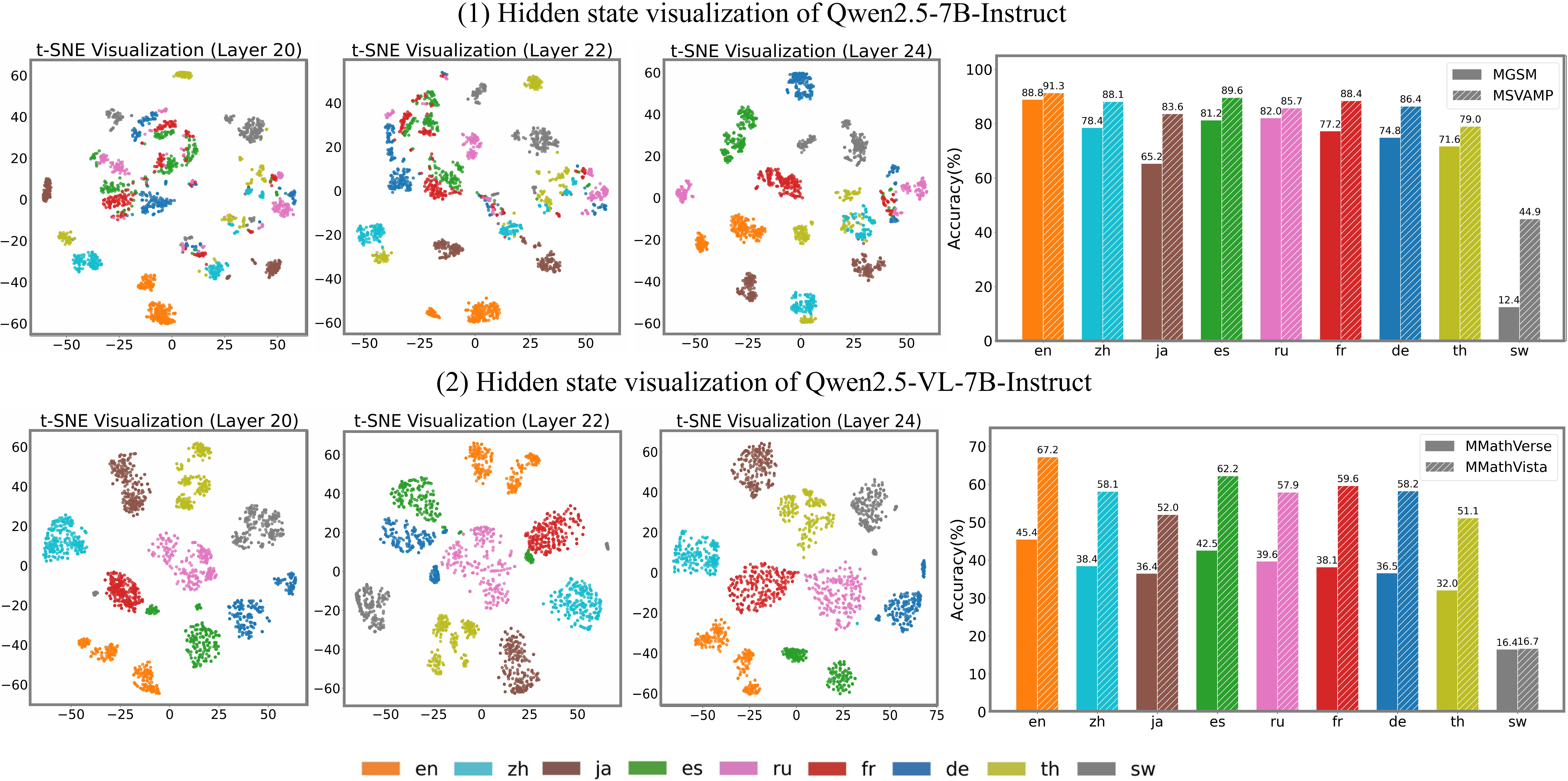}
  \caption{t-SNE hidden state visualization and reasoning performance of Qwen2.5-7B-Instruct and Qwen2.5-VL-7B-Instruct. The reasoning capability in English is substantially stronger than in other languages.}
  \label{fig:tsne}
\end{figure*}

Prior studies \cite{zhao2024large,tang2025unlocking,qin2025survey,li2025causal,li2025cai,zhao2025less} revealed the internal mechanism of multilingual reasoning: hidden states are transformed into high-resource language representations (e.g., English) in early layers, then exploited for reasoning from middle to later layers, and finally restored target language features in late layers. 
However, as shown in Figure~\ref{fig:tsne}, we observe that hidden states in reasoning-related layers of both LLMs and LVLMs still exhibit significant differences between English and non-English inputs.
This consistent cross-architecture phenomenon motivates us to steer the distribution of non-English hidden states toward English, endowing both LLMs and LVLMs with English-level reasoning capabilities under a general framework. As shown in Figure~\ref{fig:pre}, we propose a training-free inference-time method to enhance \textbf{M}ultilingual \textbf{R}easoning capabilities via \textbf{R}epresentation \textbf{E}ngineering (\textbf{MRRE}) without using any additional training data or tools. \textbf{MRRE} sequentially injects two precomputed vectors at specific layers during forward passing: \textit{\textbf{cross-lingual reasoning enhancement vectors}}, which steer non-English reasoning representations toward English distributions to strengthen reasoning, and \textbf{\textit{target-language output anchoring vectors}}, which restore the distribution of the target language to preserve input–output language consistency.

Experimental results of six advanced LLMs and LVLMs on four reasoning benchmarks demonstrate that MRRE enhances non-English reasoning by an average gain of 5.48\% and up to 7.54\% in low-resource languages (Thai \& Swahili), while improving input-output language consistency by 3.78\%.

\section{Related Works}

\subsection{Multilingual Foundation Models}

To address the multilingual demands of real-world global applications, recent studies have extended foundation models to multilingual models. 
Advanced LLMs like Qwen3-8B \cite{yang2025qwen3} and Llama-3.1-8B-Instruct \cite{grattafiori2024llama}, are pretrained and instruction-tuned on massive multilingual data, enabling them to respond across diverse languages.
Furthermore, advanced LVLMs like Qwen2.5-VL-7B-Instruct \cite{qwen2.5-VL} and InternVL3.5-8B-Chat \cite{wang2025internvl3_5}, integrate visual encoders with these language backbones, thereby inheriting comparable multilingual capabilities.
However, the unbalanced distribution of training data across languages results in a significant performance disparity between high-resource languages (e.g., English) and low-resource language (e.g., Swahili), raising concerns of fairness.

\subsection{Multilingual Reasoning Enhancement}

To enhance multilingual reasoning, previous research can be divided into two categories: \textbf{(1) Data-driven training} methods enhance multilingual reasoning by aligning cross-lingual representations or fine-tuning with multilingual supervision, including contrastive alignment \cite{li2023improving, huang2024mindmerger, arora2024towards} and reasoning-specific instruction tuning \cite{ zhang2024lingualift, geng2024not, lai2024mcot, fan2025slam}. Although effective, these approaches incur substantial data and computational costs. \textbf{(2) Prompting} methods mitigate language imbalance without parameter updates, including through direct multilingual inputs \cite{sakai2024mcsqa,khandelwal2024moral}, pivot-language translation \cite{liu2024translation}, or chain-of-thought prompting \cite{wang2024m4u,qin2023cross}, although effectiveness remain sensitive to translation quality and prompt design, accompanying by high latency. 
In contrast to previous work, MRRE is the first training-free inference-time method via representation engineering without using any additional data or tools,
which clearly distinguishes MRRE from existing methods.

\section{Methods}

\begin{figure*}[!ht]
  \includegraphics[width=1.0\linewidth]{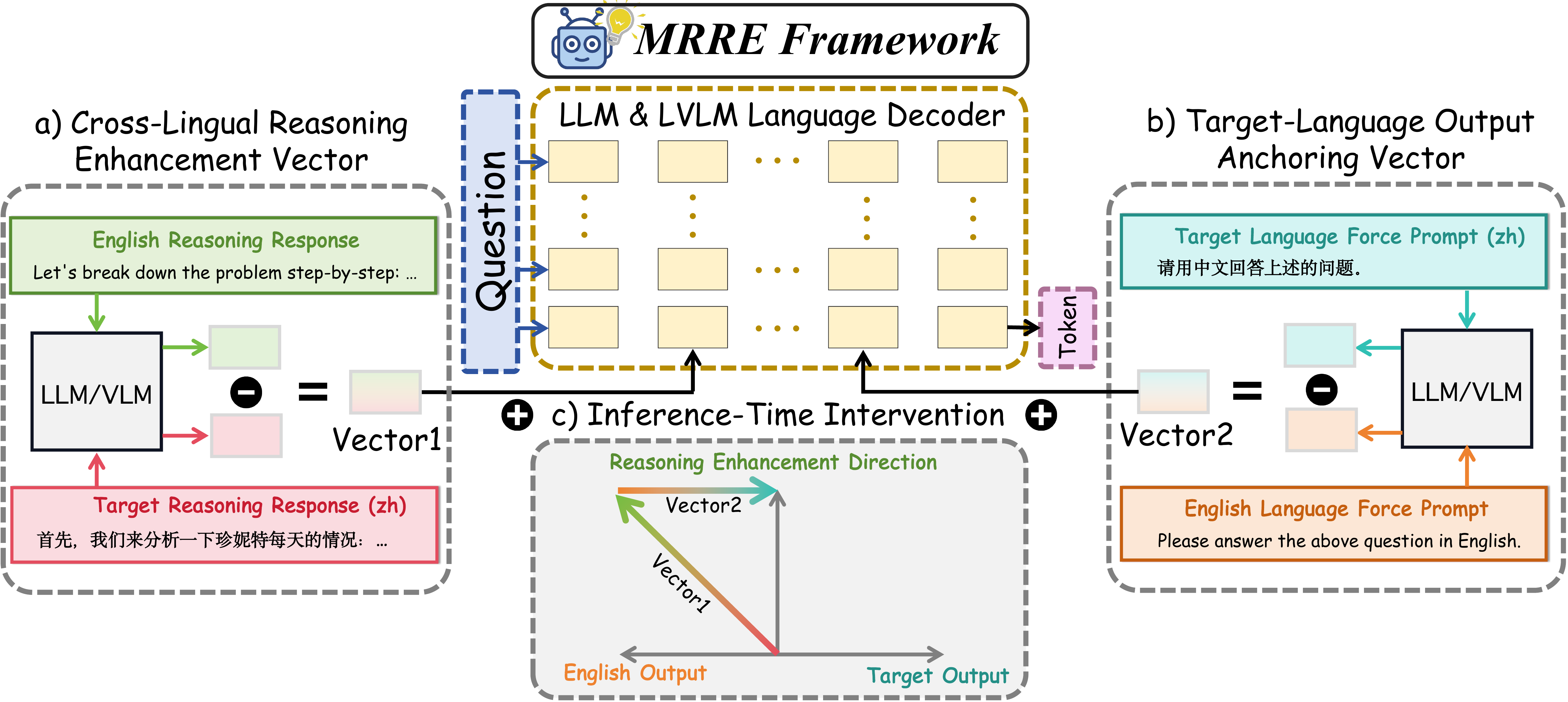}
  \caption{An overview of our proposed MRRE method. Each rectangle represents the model's hidden state during the forward passing. MRRE consists of three key stages: \textbf{{a) Cross-Lingual Reasoning Enhancement Vectors}} \S\ref{reasoning} are derived from the hidden state differences between English and non-English reasoning responses. \textbf{{b) Target-Language Output Anchoring Vectors}} \S\ref{anchoring} are derived from the hidden state differences between non-English and English language forcing prompts. \textbf{{c) Hierarchical Inference-Time Intervention}} \S\ref{intervention}: Precomputed vectors are sequentially injected into the last-token representations at specific layers during forward passing, thereby enhancing non-English reasoning capabilities while preserving input-output language consistency. }
  \label{fig:method}
\end{figure*}

\subsection{Task Formulation}
We restrict our scope to models that are based on auto-regressive Transformer architecture \cite{vaswani2017attention}, as it is adopted by most SOTA LLMs and LVLMs. The input sequence of LLMs and LVLMs are processed through $L$ transformer layers of the language decoder, each consisting of multi-head self-attention (MHSA), feed-forward network (FFN) that is usually a multilayer perception (MLP), and a residual stream is applied between each component.   
The hidden state $\boldsymbol{h}^{(l)} \in \mathbb{R}^d$ for token $t$ at layer $l$ under input sequence $R(x)$ can be computed from the previous layer:
\begin{equation}
    \boldsymbol{h}^{(l)}(R(x),t) = \boldsymbol{h}^{(l-1)}(R(x),t) + \boldsymbol{a}^{(l)} + \boldsymbol{m}^{(l)},
\label{eq:residual_stream}
\end{equation}
where $\boldsymbol{a}^{(l)}$ and $\boldsymbol{m}^{(l)}$ are the outputs of the MHSA and FFN component at layer $l$. Finally, the model predicts the next token in an auto-regressive manner based on the hidden state of the last layer.  

In this paper, to bridge the significant performance gap between English and non-English languages in reasoning tasks, we propose a representation engineering approach that applies a two-stage hierarchical intervention on the hidden states of specific layers within the language decoder, unlocking non-English reasoning capabilities while preserving input-output language consistency.

\subsection{Cross-Lingual Reasoning Enhancement}
\label{reasoning}
Prior studies \cite{zhao2024large,tang2025unlocking} have shown that the mid-layer hidden states play a critical role in shaping reasoning latent representation.
Furthermore, as shown in Figure \ref{fig:tsne}, we observe that the hidden states from middle to deeper layers under different languages exhibit significant differences in the latent space across both LLMs and LVLMs. These findings motivate us to propose a representation engineering strategy that aligns the reasoning capability of non-English with English, which exhibits stronger reasoning performance. 

Since LLMs and LVLMs generate tokens in an auto-regressive manner, we focus on the hidden state of the last token, which aggregates the most comprehensive visual and textual information. 
To precisely estimate the reasoning enhancement direction for hidden states, we define \textbf{\textit{cross-lingual reasoning enhancement vectors}}, which align the hidden states of the reasoning chain in the target non-English language with the stronger reasoning chain in English. 
These vectors are computed by comparing the output last token's hidden states on a set of reasoning problems $\mathcal{X}$. 
For each problem $x \in \mathcal{X}$, we feed the model with both English prompt and parallel non-English prompt, generating the corresponding reasoning responses, $R_\mathcal{E}(x)$ and $R_\mathcal{T}(x)$. 
We define a difference vector $\Delta\boldsymbol{h}_x^{(l)}$ to estimate the latent difference between English and non-English:
\begin{equation}
\Delta\boldsymbol{h}_x^{(l)} = \boldsymbol{h}^{(l)}(R_\mathcal{E}(x), t_{\text{last}}) - \boldsymbol{h}^{(l)}(R_\mathcal{T}(x), t_{\text{last}}),
\label{eq:reasoning_vector_diff}
\end{equation}
The \textbf{\textit{cross-lingual reasoning enhancement vector}} at $l$-th layer, $\boldsymbol{v}_r^{(l)}$, is then computed as the mean of these difference vectors over the entire set $\mathcal{X}$:
\begin{equation}
\boldsymbol{v}_r^{(l)} = \frac{1}{|\mathcal{X}|} \sum_{x \in \mathcal{X}} \Delta\boldsymbol{h}_x^{(l)}.
\label{eq:reasoning_vector}
\end{equation}

By steering the target-language hidden states along $\boldsymbol{v}_r^{(l)}$, we shift the non-English reasoning chain to a stronger reasoning chain shaped by English. This intervention enables the model to exhibit English-level reasoning capability when answering questions in the non-English language.

\subsection{Target-Language Output Anchoring}
\label{anchoring}
However, merely applying the \textit{\textbf{cross-lingual enhancement vectors}} may compromise input-output language consistency. 
When the steered hidden states pass through the layers responsible for constructing the output language representation, the models process them as if they were English inputs, resulting in English outputs rather than the target language.
To address this issue, we propose \textit{\textbf{target-language output anchoring vectors}}, which guide the model to steer the English-like output distribution toward the target non-English language in the layers responsible for constructing the output language representation, thereby ensuring input-output language consistency. 
The \textit{\textbf{target-language output anchoring vector}} at $l'$-th layer, $\boldsymbol{v}_a^{(l')}$, is computed by taking the difference between the last token representations of two fixed language forcing prompts: $P_{\mathcal{T}}$ ("Please answer this question in \textit{<target language>}" translated into \textit{<target language>}), and $P_{\mathcal{E}}$ ("Please answer this question in English"):
\begin{equation}
    \boldsymbol{v}_a^{(l')} = \boldsymbol{h}^{(l')}(P_{\mathcal{T}}, t_{\text{last}}) - \boldsymbol{h}^{(l')}(P_{\mathcal{E}}, t_{\text{last}}).
\label{eq:anchoring_vector}
\end{equation}
This vector provides a precise estimate of the hidden state distribution shift from English to the target language, enabling the model to recover output target-language distribution in later layers and ensuring input-output language consistency.

\subsection{Hierarchical Inference-Time Intervention}
\label{intervention}

Considering the autoregressive decoding mechanism of language models, we propose a two-stage, hierarchical inference-time intervention method. First we apply \textit{\textbf{cross-lingual enhancement vectors}} to the last token of the current reasoning response at middle layer, steering non-English hidden state $\boldsymbol{h}_t^{(l)}$ to an English-like hidden state $\tilde{\boldsymbol{h}}^{(l)}$:
\begin{equation}
    \hat{\boldsymbol{h}}^{(l)} = \boldsymbol{h}^{(l)} + \alpha_1 \cdot \tilde{\boldsymbol{v}}_r^{(l)}, \tilde{\boldsymbol{h}}^{(l)} = \hat{\boldsymbol{h}}^{(l)} \cdot \frac{\lVert \boldsymbol{h}^{(l)} \rVert_2}{\lVert \hat{\boldsymbol{h}}^{(l)} \rVert_2},
\label{eq:intervention_1}
\end{equation}
where $\alpha_1$ denotes scaling coefficients and $|| \cdot ||$ represents the $\ell_2$ norms of the activation vectors. The normalization strategy ensures that the vector scale remains consistent before and after intervention, preventing undesired magnitude shifts that may distort downstream representations.

The model exhibits English-level reasoning capability from middle to later layers as if processing an English problem, producing English output hidden states ${\boldsymbol{h}}^{(l')}$ at $l'$-th layer. Then we apply the \textit{\textbf{target-language output anchoring vectors}} to steer English output hidden state back to the target $\tilde{\boldsymbol{h}}^{(l')}$:
\begin{equation}
    \hat{\boldsymbol{h}}^{(l')} = \boldsymbol{h}^{(l')} + \alpha_2 \cdot \tilde{\boldsymbol{v}}_a^{(l')}, \tilde{\boldsymbol{h}}^{(l')} = \hat{\boldsymbol{h}}^{(l')} \cdot \frac{\lVert \boldsymbol{h}^{(l')} \rVert_2}{\lVert \hat{\boldsymbol{h}}^{(l')} \rVert_2},
\label{eq:intervention_2}
\end{equation}

Finally, the newly generated token is then appended to the current input sequence. In the middle layers, the hidden state of updated input sequence can be \textit{“translated”} into English, activating stronger reasoning capabilities under English-like states; and in later layers, it can be \textit{“back-translated”} to the target language, ensuring input-output language consistency.
This information flow continues until end-of-sequence token is produced.

\begin{table*}[!ht]
\centering
\small
\renewcommand{\arraystretch}{0.80}
\setlength{\tabcolsep}{0.01pt}

\begin{tabular*}{\textwidth}{l  @{\extracolsep{\fill}} *{10}{S[table-format=2.1]}  *{10}{S[table-format=2.1]} }
\toprule

\multicolumn{1}{c}{\textbf{Model}} & {\textbf{En}} & {\textbf{Zh}} & {\textbf{Ja}} & {\textbf{Es}} & {\textbf{Ru}} & {\textbf{Fr}} & {\textbf{De}} & {\textbf{Th}} & {\textbf{Sw}} &\textbf{LC}& {\textbf{En}} & {\textbf{Zh}} & {\textbf{Ja}} & {\textbf{Es}} & {\textbf{Ru}} & {\textbf{Fr}} & {\textbf{De}} & {\textbf{Th}} & {\textbf{Sw}} &\textbf{LC}  \\

\midrule

\multicolumn{1}{c}{\textbf{\textit{LLMs}}} & \multicolumn{10}{c}{\textbf{\textit{MGSM}}} & \multicolumn{10}{c}{\textbf{\textit{MSVAMP}}} \\
\cmidrule(lr){1-1} \cmidrule(lr){2-11} \cmidrule(lr){12-21}
\rowcolor{gray!15} Qwen2.5-7B-Instruct & 88.8 & 78.4 & 65.2 & \cellcolor{green!15}\textbf{81.2} & 82.0 & 77.2 & 74.8 & 71.6 & 12.4 &{84.7}& 91.3 & 88.1 & 83.6 & 89.6 & 85.7 & 88.4 & 86.4 & 79.0 & 44.9&86.3 \\
\quad + Few-Shot 
& {--} & 79.1 & 66.8 & 80.2 & 82.5 & 77.8 & 75.6 & 72.3 & 14.9 & 87.6
& {--} & 88.6 & 84.1 & \cellcolor{green!15}\textbf{90.1} & 86.2 & 89.0 & 87.1 & 80.4 & 46.8 & 88.9 \\

\quad + SFT
& {--} & 76.4 & 64.9 & 78.6 & 80.8 & 75.9 & 74.2 & 70.1 & 13.7 & 86.0
& {--} & 86.9 & 82.3 & 88.7 & 84.6 & 87.5 & 85.9 & 78.8 & 45.1 & 87.4 \\

\quad + Multilingual-CoT
& {--} & 80.6 & 68.9 & 80.4 & 83.9 & 78.6 & 77.3 & 73.9 & 15.3 & 82.1
& {--} & 88.4 & 85.2 & 91.3 & 86.5 & 90.0 & \cellcolor{green!15}\textbf{88.2} & 81.6 & 49.3 & 84.8 \\

\quad + Language forcing &{--} &77.2 &67.6 &79.2 &82.0 &76.0 &76.8 &71.6 & 16.0 &92.2&{--} & 86.2 &82.0 &88.8 &85.1 &87.8 &86.4 & 79.6 & 45.7 &89.4\\
\quad + MRRE & {--} & \cellcolor{green!15}\textbf{81.6} & \cellcolor{green!15}\textbf{71.2} & \cellcolor{green!15}\textbf{81.2} & \cellcolor{green!15}\textbf{84.7} & \cellcolor{green!15}\textbf{79.2} & \cellcolor{green!15}\textbf{78.4} & \cellcolor{green!15}\textbf{74.8} & \cellcolor{green!15}\textbf{17.2} &\cellcolor{green!15}\textbf{92.7} 
&{--} & \cellcolor{green!15}\textbf{88.7} & \cellcolor{green!15}\textbf{84.7} & 90.0 & \cellcolor{green!15}\textbf{86.5} & \cellcolor{green!15}\textbf{90.1} & 87.4 &\cellcolor{green!15} \textbf{82.6} & \cellcolor{green!15}\textbf{52.3} &\cellcolor{green!15}\textbf{90.3} \\
\midrule
\rowcolor{gray!15} Qwen3-8B & 90.8 & 80.4 & 76.4 & 83.2 & 88.4 & 82.4 & 84.0 & 85.6 & 40.4 &89.3& 91.9 & 88.1 & 87.4 & 90.7 & 87.0 & 90.1 & 89.1 & 83.1 & 67.8 &89.4 \\
\quad + Language forcing &{--} &82.0 &76.8 &84.0 &\cellcolor{green!15}\textbf{88.8} &79.6 &82.0 &84.0 &37.6 &93.2& {--} & 88.6&86.8&90.0&87.7&89.5&88.7&83.1&62.3&91.8\\
\quad + MRRE &{--} &\cellcolor{green!15}\textbf{84.7} &\cellcolor{green!15}\textbf{81.4} &\cellcolor{green!15}\textbf{84.2} &\cellcolor{green!15}\textbf{88.8} &\cellcolor{green!15}\textbf{84.1} &\cellcolor{green!15}\textbf{85.5} &\cellcolor{green!15}\textbf{86.0} &\cellcolor{green!15}\textbf{45.8} &\cellcolor{green!15}\textbf{95.6}&{--} &\cellcolor{green!15}\textbf{89.1}&\cellcolor{green!15}\textbf{88.2} &\cellcolor{green!15}\textbf{90.9}&\cellcolor{green!15}\textbf{88.3}&\cellcolor{green!15}\textbf{90.3}&\cellcolor{green!15}\textbf{89.4}&\cellcolor{green!15}\textbf{85.4}&\cellcolor{green!15}\textbf{73.4}&\cellcolor{green!15}\textbf{92.5} \\
\midrule
\rowcolor{gray!15} Llama-3.1-8B- Instruct & 77.2 & 61.2 & 38.0 & 64.8 & 53.2 & 63.6 & 67.2 & 54.4 & 44.8 &95.3 &78.4 & 65.7 & 53.8 & 71.4 & 56.6 & 76.4 & 72.9 & 55.8 & 50.8 &94.9\\
\quad + Language forcing &{--} &57.6 &44.8 &63.6 &62.4 &68.0 &66.4 &53.2 & 45.6&95.1&{--} &58.5 &59.0&71.50&\cellcolor{green!15}\textbf{66.0}&73.0&69.90&56.50&50.5 &95.1 \\
\quad + MRRE &{--} & \cellcolor{green!15}\cellcolor{green!15}\textbf{66.6} &\cellcolor{green!15}\cellcolor{green!15}\textbf{47.7} &\cellcolor{green!15}\cellcolor{green!15}\textbf{68.7} &\cellcolor{green!15}\cellcolor{green!15}\textbf{64.5} &\cellcolor{green!15}\cellcolor{green!15}\textbf{70.0} &\cellcolor{green!15}\cellcolor{green!15}\textbf{71.2} &\cellcolor{green!15}\cellcolor{green!15}\textbf{64.3} &\cellcolor{green!15}\cellcolor{green!15}\textbf{53.9} &\cellcolor{green!15}\textbf{96.6}& {--} & \cellcolor{green!15}\textbf{70.6} &\cellcolor{green!15}\textbf{64.3} &\cellcolor{green!15}\textbf{73.2} &65.4 &\cellcolor{green!15}\textbf{76.4}&\cellcolor{green!15}\textbf{74.2}&\cellcolor{green!15}\textbf{63.9}&\cellcolor{green!15}\textbf{57.6}&\cellcolor{green!15}\textbf{96.4} \\
\midrule
\multicolumn{1}{c}{\textbf{\textit{LVLMs}}} & \multicolumn{10}{c}{\textbf{\textit{MMathVerse}}} & \multicolumn{10}{c}{\textbf{\textit{MMathVista}}} \\
\cmidrule(lr){1-1} \cmidrule(lr){2-11} \cmidrule(lr){12-21}
\rowcolor{gray!15} Qwen2.5-VL-7B-Instruct & 45.4 &38.4 &36.4 &42.5 &39.6 & 38.1&36.5 &32.0&16.4 &92.5 &67.2&58.1&52.0&62.2&57.9&59.6&58.2&51.1&16.7&94.1\\
\quad + Few-Shot & {--} &45.3 &40.0&45.2&44.1&42.7&43.1&35.5&23.0 & 92.6 &{--}&56.9&57.4&60.9&59.7&61.6&56.9&52.7&34.8&94.0 \\
\quad + SFT &{--}&34.0&33.4&33.9&34.5&30.4&31.7&28.3&17.0&92.7&{--}& 51.5& 49.1&54.3&54.7&53.9&51.7&44.9&18.2&94.5 \\
\quad + Multilingual-CoT &{--}&44.9&42.0&46.7&44.5&44.9&44.0&37.8&22.4&90.4&{--} & 61.7&57.4&63.9&62.5& 59.7&61.2&52.8&30.3&92.6\\
\quad + Language forcing &{--} &36.8 &37.7 &43.6 &40.6 &39.6 &40.0 &43.5 &17.8 &91.8&{--}& 60.0&53.0&61.2&58.3&59.4&56.9&53.7&10.1&94.4  \\
\quad + MRRE & {--} &\cellcolor{green!15}\textbf{45.9} & \cellcolor{green!15}\textbf{44.6}& \cellcolor{green!15}\textbf{47.5}&\cellcolor{green!15}\textbf{44.6} &\cellcolor{green!15}\textbf{45.8} &\cellcolor{green!15}\textbf{45.7} & \cellcolor{green!15}\textbf{48.4}&\cellcolor{green!15}\textbf{33.8} &\cellcolor{green!15}\textbf{93.1}&{--} &\cellcolor{green!15}\textbf{62.3}&\cellcolor{green!15}\textbf{61.1}&\cellcolor{green!15}\textbf{64.2}&\cellcolor{green!15}\textbf{62.8}&\cellcolor{green!15}\textbf{61.2}&\cellcolor{green!15}\cellcolor{green!15}\textbf{64.1}  &\cellcolor{green!15}\textbf{57.3}&\cellcolor{green!15}\textbf{28.6}&\cellcolor{green!15}\textbf{95.5}\\
\midrule
\rowcolor{gray!15} LLaVA-Onevision-7B &29.9&28.9&25.8&29.2&26.2&26.8&25.3&23.5&11.3&76.9&61.9&51.1&43.7&\cellcolor{green!15}\textbf{58.0}&49.2&55.8&48.0&41.7&16.3&75.0\\
\quad + Language forcing &{--} &26.6&22.7&25.9&22.9&20.3&22.0&26.3&4.0&\cellcolor{green!15}\textbf{77.8}&{--}&44.6 &38.6 &45.1&41.1&41.8 &37.1 &26.7 &10.8&78.4  \\
\quad + MRRE &{--} &\cellcolor{green!15}\textbf{32.9} &\cellcolor{green!15}\textbf{30.0} &\cellcolor{green!15}\textbf{30.0} &\cellcolor{green!15}\textbf{28.2} &\cellcolor{green!15}\textbf{30.0} &\cellcolor{green!15}\textbf{29.5} &\cellcolor{green!15}\textbf{26.5} &\cellcolor{green!15}\textbf{19.1} &\cellcolor{green!15}\textbf{77.8}&{--}&\cellcolor{green!15}\textbf{52.3}&\cellcolor{green!15}\textbf{48.2}&57.0&\cellcolor{green!15}\textbf{55.7}&\cellcolor{green!15}\textbf{56.6}&\cellcolor{green!15}\textbf{54.7}&\cellcolor{green!15}\textbf{48.3}&\cellcolor{green!15}\textbf{32.1} &\cellcolor{green!15}\textbf{80.3}\\
\midrule
\rowcolor{gray!15} InternVL3.5-VL-8B-Chat &57.6&52.3&48.9&52.3&49.0&48.5&46.6&34.4&32.3&76.3&72.2&59.7&60.2&67.9&68.7&67.9&68.7&62.5&40.6&79.6\\
\quad + Language forcing &{--}&54.7&50.8&51.4&49.4&49.1&49.8&35.5&36.3&\cellcolor{green!15}\textbf{84.2}&{--} &62.4 &61.1 &64.1 &62.9 &61.5 &64.1 &57.8 &34.8&85.6   \\
\quad + MRRE &{-} &\cellcolor{green!15}\textbf{55.6} &\cellcolor{green!15}\textbf{51.2} &\cellcolor{green!15}\textbf{53.3} &\cellcolor{green!15}\textbf{50.1} &\cellcolor{green!15}\textbf{50.3} &\cellcolor{green!15}\textbf{52.3}&\cellcolor{green!15}\textbf{44.6} &\cellcolor{green!15}\textbf{42.3} &81.9&{--}&\cellcolor{green!15}\textbf{63.5}&\cellcolor{green!15}\textbf{66.9}&\cellcolor{green!15}\textbf{70.7}&\cellcolor{green!15}\textbf{70.5}&\cellcolor{green!15}\textbf{70.1}&\cellcolor{green!15}\textbf{69.2}&\cellcolor{green!15}\textbf{66.3}&\cellcolor{green!15}\textbf{46.6} & \cellcolor{green!15}\textbf{86.4} \\
\bottomrule
\end{tabular*}
\caption{Accuracy (\%) and Language Consistency (LC, \%) of three advanced LLMs and three advanced LVLMs with different settings across 8 languages (Zh, Ja, Es, Ru, Fr, De, Th, Sw) and 4 reasoning benchmarks: \textit{MGSM, MSVAMP, MMathVerse, MMathVista}. Best performances for each experimental settings are \textbf{bolded}.}
\label{main1}
\end{table*}

\begin{table*}[!ht]
\renewcommand{\arraystretch}{0.80}
\setlength{\tabcolsep}{0.5pt}
\centering
\small
\sisetup{detect-weight, mode=text}
\renewrobustcmd{\bfseries}{\fontseries{b}\selectfont}

\begin{tabular}{llllllllll}
\toprule
\multicolumn{1}{c}{\multirow{2}{*}{\textbf{Model}} }& \multicolumn{6}{c}{\textbf{\textit{MMathVerse}}} & \multicolumn{3}{c}{\textbf{\textit{MMathVista}}} \\
\cmidrule(lr){2-7} \cmidrule(lr){8-10}
 & {\textbf{T-D}} & {\textbf{T-L}} & {\textbf{V-I}} & {\textbf{V-D}} & {\textbf{V-O}} & {\textbf{Overall}} & {\textbf{General}} & {\textbf{Math}} & {\textbf{Overall}}  \\
\midrule
\rowcolor{gray!15}
Qwen2.5-VL-7B &39.9 &35.1&32.1&32.5&34.6&34.9&53.5&50.7&52.0 \\
\quad + Few-Shot & 46.2 &40.4& 37.8& 37.2& 37.8&39.9& 55.0& 55.2& 55.1  \\
\quad + SFT & 35.6 & 31.6 & 28.5 &28.2& 28.1 &30.4&42.8&48.0&45.8 \\
\quad + Multilingual-CoT &47.3 & 41.6 &39.9&39.9& 40.3& 41.8&57.0&55.5&56.2  \\
\quad + Language forcing &44.0&37.2&34.9&35.1&37.0&37.7&52.6&50.8&51.6 \\
\quad + MRRE 
 & \cellcolor{green!15}\textbf{51.2}\textcolor{green!40!black}{(+11.3)} 
 & \cellcolor{green!15}\textbf{45.8}\textcolor{green!40!black}{(+10.7)} 
 & \cellcolor{green!15}\textbf{42.2}\textcolor{green!40!black}{(+10.1)} 
 & \cellcolor{green!15}\textbf{41.2}\textcolor{green!40!black}{(+8.7)} 
 & \cellcolor{green!15}\textbf{41.9}\textcolor{green!40!black}{(+7.3)} 
 & \cellcolor{green!15}\textbf{44.5}\textcolor{green!40!black}{(+9.6)} 
 & \cellcolor{green!15}\textbf{55.5}\textcolor{green!40!black}{(+2.0)} 
 & \cellcolor{green!15}\textbf{58.5}\textcolor{green!40!black}{(+7.8)} 
 & \cellcolor{green!15}\textbf{57.1}\textcolor{green!40!black}{(+5.1)} \\
\midrule
\rowcolor{gray!15}
LLaVA-Onevision-7B &27.5&28.4&24.8&22.6&\cellcolor{green!15}\textbf{22.4}&24.6&44.7&46.1&45.5  \\
\quad + Language forcing &28.4&25.7&24.2&19.8&8.5&29.3&21.3&41.0&35.8  \\
\quad + MRRE 
 & \cellcolor{green!15}\textbf{37.4}\textcolor{green!40!black}{(+9.9)} 
 & \cellcolor{green!15}\textbf{32.1}\textcolor{green!40!black}{(+3.7)} 
 & \cellcolor{green!15}\textbf{30.9}\textcolor{green!40!black}{(+6.1)} 
 & \cellcolor{green!15}\textbf{28.3}\textcolor{green!40!black}{(+5.7)} 
 & \cellcolor{red!15}21.2\textcolor{red!75!black}{(-1.2)} 
 & \cellcolor{green!15}\textbf{30.0}\textcolor{green!40!black}{(+5.4)} 
 & \cellcolor{green!15}\textbf{48.6}\textcolor{green!40!black}{(+3.9)} 
 & \cellcolor{green!15}\textbf{52.3}\textcolor{green!40!black}{(+6.2)} 
 & \cellcolor{green!15}\textbf{50.6}\textcolor{green!40!black}{(+5.1)} \\
\midrule
\rowcolor{gray!15}
InternVL3.5-8B &48.3&41.4&33.3&39.4&37.5&40.1&62.0&61.0&61.5  \\
\quad + Language forcing &47.8&40.3&36.1&38.1&41.7&40.8&60.8&61.8&61.4  \\
\quad + MRRE 
 & \cellcolor{green!15}\textbf{53.0}\textcolor{green!40!black}{(+4.7)} 
 & \cellcolor{green!15}\textbf{47.2}\textcolor{green!40!black}{(+5.8)} 
 & \cellcolor{green!15}\textbf{41.7}\textcolor{green!40!black}{(+8.4)} 
 & \cellcolor{green!15}\textbf{44.3}\textcolor{green!40!black}{(+4.9)} 
 & \cellcolor{green!15}\textbf{48.0}\textcolor{green!40!black}{(+10.5)} 
 & \cellcolor{green!15}\textbf{46.8}\textcolor{green!40!black}{(+6.7)} 
 & \cellcolor{green!15}\textbf{65.3}\textcolor{green!40!black}{(+3.3)} 
 & \cellcolor{green!15}\textbf{64.0}\textcolor{green!40!black}{(+3.0)} 
 & \cellcolor{green!15}\textbf{63.8}\textcolor{green!40!black}{(+2.3)} \\
\bottomrule
\end{tabular}
\caption{Mean performance (\%) of three advanced LVLMs with diverse experimental settings across 8 languages (Zh, Ja, Es, Ru, Fr, De, Th, Sw) and 2 benchmarks: MMathVerse (\textit{Text-Dominant, Text-Lite, Vision-Integrated, Vision-Dominant, and Vision-Only categories, Overall}), MathVista (\textit{General, and Math-related categories, Overall}). 
}
\label{main2}
\end{table*}

\section{Experiments}
\subsection{Experimental Setup}
To thoroughly evaluate our proposed MRRE's performance and generalization across models and languages, we construct systematic experiments.

\noindent\textbf{Baseline Models.} We evaluate our proposed MRRE method on six SOTA models to demonstrate its broad applicability across LLM and LVLMs.

\textbf{LLMs:} Qwen2.5-7B-Instruct \cite{qwen2.5}, Qwen3-8B \cite{yang2025qwen3}, and Llama-3.1-8B-Instruct \cite{grattafiori2024llama}.

\textbf{LVLMs:} Qwen2.5-VL-7B-Instruct \cite{qwen2.5-VL}, LLaVA-OneVision \cite{li2024llavaonevisioneasyvisualtask} and InternVL3.5-8B-Chat \cite{wang2025internvl3_5}.

\noindent\textbf{Baseline Methods.} We select several methods that are effective for multilingual reasoning tasks in both LLMs and VLMs, including Few-Shot, SFT, Multilingual-CoT \cite{barua2025long}, and Language forcing \cite{wang2025polymath}.

\noindent\textbf{Benchmarks.} We select four challenging benchmarks to evaluate multilingual reasoning of both LLMs and LVLMs, covering both mathematical and general reasoning capabilities, including \textbf{MGSM} \cite{shi2022language}, \textbf{MSVAMP} \cite{chen2023breaking}, \textbf{MMathVerse} \cite{zhang2024mathverse} and \textbf{MMathVista} \cite{lu2023mathvista}. 

We adopt \textit{Accuracy} as the evaluation metric. Language Consistency (LC) denotes the proportion of responses generated in the target language.

\paragraph{Implementation Details.} We randomly sample 100 instances from MGSM and MathVerse to construct MRRE vectors for LLMs and LVLMs, respectively. All experiments are conducted on 8 $\times$ NVIDIA A100 80GB. See Appendix \ref{exidetails} for details.

\subsection{Main Results}
Based on experimental results presented in Table \ref{main1} \& \ref{main2}, we can draw the following key conclusions:
\paragraph{(1) Effective multilingual reasoning enhancement performance.} MRRE achieves effective performance across high-resource and low-resource non-English languages on LLMs and LVLMs, leading to an average improvement of 5.48\% on four benchmarks, \textbf{while improving input-output language consistency by 3.78\%.} Notably, improvements are significantly enhanced in low-resource languages (Th, Sw) by 7.54\%. 
\paragraph{(2) Multimodal generalizability.} MRRE consistently improves performance across purely textual and vision–language reasoning tasks. Notably, results on MMathVerse demonstrate that MRRE enhances reasoning in all five categories, including \textit{Text-Dominant}, \textit{Text-Lite}, \textit{Vision-Integrated}, \textit{Vision-Dominant}, and \textit{Vision-Only} tasks, proving MRRE's effectiveness regardless of modality composition.
\paragraph{(3) Cross-model generalizability.} By systematically evaluating diverse LLM and LVLM backbones, we find that MRRE does not rely on a particular model architecture. In contrast, MRRE benefits from universal cross-lingual difference in hidden states, ensuring broad application across other SOTA open-source models.
\paragraph{(4) Cross-dataset generalizability.} Although the construction of MRRE vectors relies on reasoning data from MGSM and MMathVerse, these vectors remain effective on out-of-distribution datasets such as MSVAMP and MMathVista, 
suggesting that the intervention represents a generalizable direction rather than being merely tailored to a specific dataset. Moreover, the improvement on MMathVista \textit{General} subset indicates that MRRE not only enhances mathematical reasoning, but also strengthens general reasoning capabilities.

\section{Analysis and Discussions}

\subsection{Analysis of Intervention}

\begin{figure}[!ht]
  \includegraphics[width=1.0\linewidth]{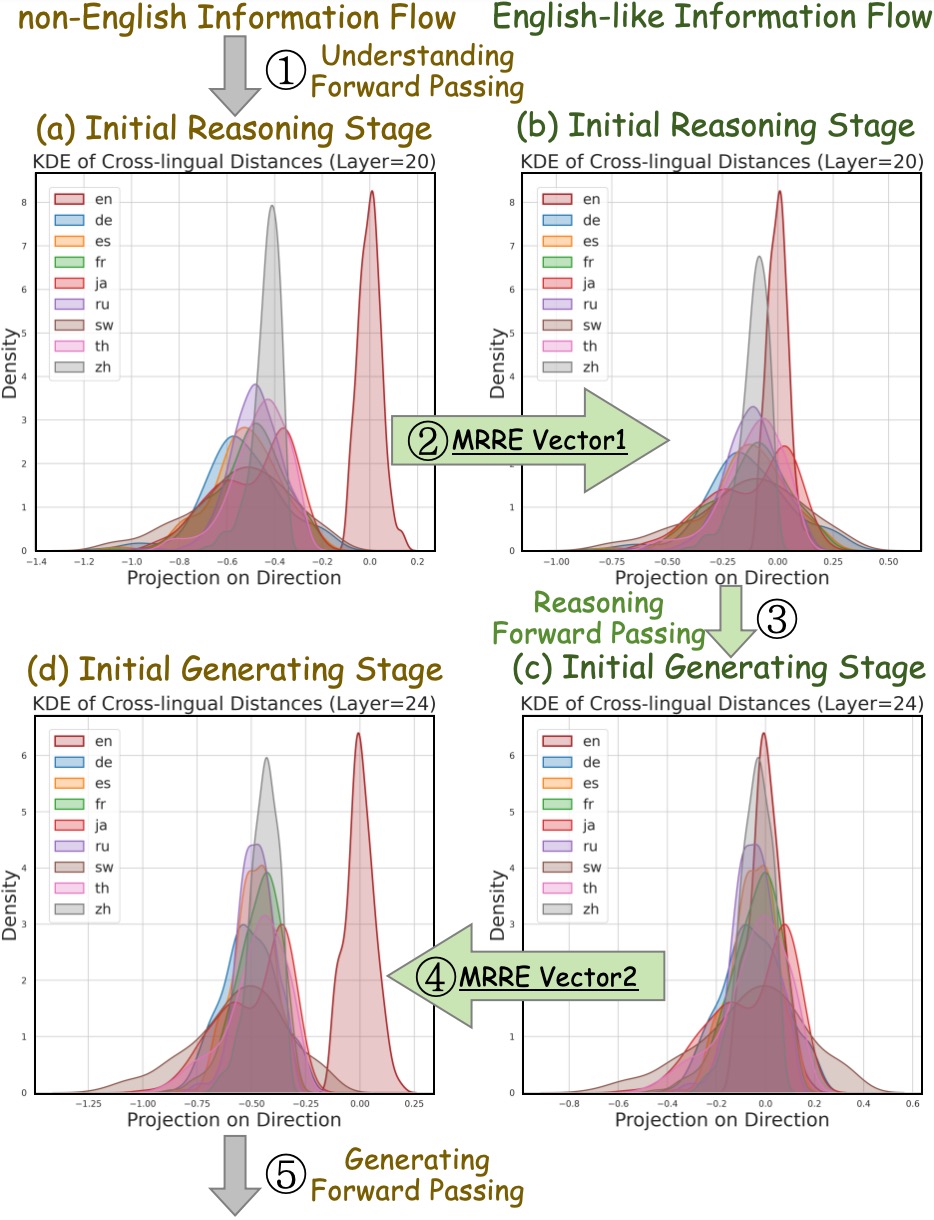}
  \caption{Kernel Density Estimate (KDE) visualization plots of cross-lingual hidden states within Qwen2.5-7B-Instruct before and after two types of intervention. The x-axis represents the SVM-derived signed distance to the mean English representation; and the y-axis represents the estimated probability density. }
  \label{hidden}
\end{figure}

To thoroughly visualize the effect of MRRE on hidden states, we conduct a stepwise analysis along the information flow of token generation: \circled{1} the model first receives non-English input and undergoes understanding forward passing until the initial reasoning stage. As shown in Figure~\ref{hidden} (a), we observe a significant difference between English and non-English representations in the latent space. Moreover, since Qwen2.5-7B-Instruct is trained on large amounts of English and Chinese data, the hidden states of these two high-resource languages exhibit much more densely clustered than others. \circled{2} Under the intervention of the \textbf{\textit{cross-lingual enhancement vectors}}, as shown in Figure \ref{hidden} (b) non-English hidden states shift closer to English. \circled{3} The English-like states then undergo reasoning forward passing until initial generating stage critical for language outputs. At this stage, the model reasons as if it were processing an English query, ultimately producing English-like representations. As illustrated in Figure \ref{hidden} (c), English and non-English hidden states become highly aligned. \circled{4} Under the intervention of the \textbf{\textit{target-language output anchoring vectors}}, as shown in Figure \ref{hidden} (d), the hidden states restore the standard output distribution of the target language, ensuring input and output language consistency. \circled{5} Finally, the anchored hidden states continue forward passing until generate the next token in target language. 
This lifetime visualization of token generation demonstrates that MRRE achieves the foundational motivation we set forth.

\subsection{Cross-modal Generalization of Vectors}
In this subsection, we thoroughly explore the cross-modal generalization of MRRE vectors.
LVLMs are trained by a \textbf{L}anguage \textbf{B}ackbone (\textit{L-B}) LLM jointly with a specific visual encoder, which results in high similarity between their output hidden states in the latent space. 
Furthermore, current language backbone LLMs exhibit stronger reasoning capabilities than corresponding LVLMs \cite{chen2025bring}.
Building on these two observations, we hypothesize that cross-lingual reasoning enhancement vectors derived from language backbones can also enhance reasoning capability of LVLMs.
To validate this hypothesis, we conduct experiments with Qwen2.5-VL-7B-Instruct and InternVL3.5-8B, replacing the cross-lingual reasoning enhancement vectors with those vectors derived from their language backbone, Qwen2.5-7B-Instruct and Qwen3-8B. As shown in Table \ref{crossmodal}, MRRE \textit{L-B} consistently improves the reasoning performance across all non-English languages. Remarkably, MRRE \textit{L-B} even surpasses MRRE \textit{Vanilla} in French and Swahili. 
These findings indicate that vectors from language backbones generalize well to LVLMs, highlighting MRRE's cross-modal generalization capability. They also suggest that LVLMs inherit multilingual reasoning capabilities from their language backbones LLMs to some extent, leading to similar multilingual shift directions.  

\begin{table}
\setlength{\tabcolsep}{1pt}
\centering
\small
\begin{tabular}{l S S S S S S S S}
\toprule
\multicolumn{1}{c}{\textbf{Model}}  & {\textbf{Zh}} & {\textbf{Ja}} & {\textbf{Es}} & {\textbf{Ru}} & {\textbf{Fr}} & {\textbf{De}} & {\textbf{Th}} & {\textbf{Sw}} \\
\midrule
\multicolumn{9}{c}{Qwen2.5-7B as \textit{Language Backbone}}\\
\midrule
\rowcolor{gray!15} Qwen2.5-VL-7B &58.1&52.0&62.2&57.9&59.6&58.2&51.1&16.7 \\
\quad+ MRRE \textit{Vanilla}  &\cellcolor{green!15}\textbf{62.3} &\cellcolor{green!15}\textbf{61.1} &\cellcolor{green!15}\textbf{64.2} &\cellcolor{green!15}\textbf{62.8} &61.2 &\cellcolor{green!15}\textbf{64.1} & \cellcolor{green!15}\textbf{57.3} &28.6  \\
\quad+ MRRE \textit{L-B} &58.9 &55.7 &62.0 &59.6 &\cellcolor{green!15}\textbf{61.4} &62.7 &54.9 &\cellcolor{green!15}\textbf{33.1}  \\
\midrule
\multicolumn{9}{c}{Qwen3-8B as \textit{Language Backbone}}\\
\midrule
\rowcolor{gray!15} InternVL3.5-8B &59.7&60.2&67.9&68.7&67.9&68.7&62.5&40.6 \\
\quad+ MRRE \textit{Vanilla} &\cellcolor{green!15}\textbf{63.5}&\cellcolor{green!15}\textbf{66.9}&\cellcolor{green!15}\textbf{70.7}&\cellcolor{green!15}\textbf{70.5}&\cellcolor{green!15}\textbf{70.1}&69.2&\cellcolor{green!15}\textbf{66.3}&46.6  \\
\quad+ MRRE \textit{L-B} &61.2&62.1&68.5&69.0&68.8&\cellcolor{green!15}\textbf{69.4}&64.5&\cellcolor{green!15}\textbf{49.4} \\
\bottomrule
\end{tabular}
\caption{Performance (\%) of the MathVista benchmark across languages. \textit{Vanilla} represents standard MRRE method, and \textit{L-B} represents MRRE method using vectors from responding LVLM's \textit{Language Backbone}.
}
\label{crossmodal}
\end{table}

\subsection{Mitigating English Bias in Latent Space}

\begin{table}
\setlength{\tabcolsep}{1pt}
\centering
\small

\label{tab:mmstar_results}
\begin{tabular}{l S S S S S S S S}
\toprule
\multicolumn{1}{c}{\textbf{Model}} & {\textbf{Zh}} & {\textbf{Ja}} & {\textbf{Es}} & {\textbf{Ru}} & {\textbf{Fr}} & {\textbf{De}} & {\textbf{Th}} & {\textbf{Sw}} \\
\midrule
\multicolumn{9}{c}{\textit{LLM on MGSM}}\\
\midrule
\rowcolor{gray!15} Qwen2.5-7B & 78.4 & 65.2 & \cellcolor{green!15}\textbf{81.2} & 82.0 & 77.2 & 74.8 & 71.6 & 12.4  \\
\quad + MRRE \textit{Vanilla} &81.6 &\cellcolor{green!15}\textbf{71.2} &\cellcolor{green!15}\textbf{81.2} &\cellcolor{green!15}\textbf{84.7} &\cellcolor{green!15}\textbf{79.2} &\cellcolor{green!15}\textbf{78.4}&74.8&17.2   \\
\quad + MRRE \textit{Debias} &\cellcolor{green!15}\textbf{83.2} &68.8 &\cellcolor{green!15}\textbf{81.2} &83.6 &77.2 &77.2 &\cellcolor{green!15}\textbf{76.4} &\cellcolor{green!15}\textbf{19.2} \\
\midrule
\multicolumn{9}{c}{\textit{LVLM on MMathVista}}\\
\midrule
\rowcolor{gray!15} Qwen2.5-VL-7B &58.1&52.0&62.2&57.9&59.6&58.2&51.1&16.7 \\
\quad+ MRRE \textit{Vanilla}  &\cellcolor{green!15}\textbf{62.3} &\cellcolor{green!15}\textbf{61.1} &\cellcolor{green!15}\textbf{64.2} &\cellcolor{green!15}\textbf{62.8} &\cellcolor{green!15}\textbf{61.2} &\cellcolor{green!15}\textbf{64.1} & \cellcolor{green!15}\textbf{57.3} &\cellcolor{green!15}\textbf{28.6}  \\
\quad + MRRE \textit{Debias}&\cellcolor{red!15}54.3 &\cellcolor{red!15}48.0 &\cellcolor{red!15}56.6 &\cellcolor{red!15}54.4 &\cellcolor{red!15}54.8 &\cellcolor{red!15}56.1&\cellcolor{red!15}48.5 &\cellcolor{red!15}11.9  \\
\bottomrule
\end{tabular}
\caption{Performance (\%) of the MGSM and MMathVista benchmark across languages. \textit{Vanilla} represents standard MRRE method, and \textit{Debias} represents MRRE method using \textbf{\textit{latent english debiasing vectors}}.
}
\label{debiasing}
\end{table}

In this subsection, we further explore how to enhance multilingual reasoning capabilities during the understanding forward passing. 
Prior study \cite{zhao2024large, xiao2026not} indicates that hidden states encode low-level semantic representations and exhibit strong linguistic characteristics at this stage.
Since a substantial portion of the training data is in English, when the model processes non-English queries, the hidden-state distribution during forward passing tends to shift toward the default English distribution. We hypothesize these shifts in early layers may distort the encoded low-level semantic representations, leading to performance drops.
To address this issue, we introduce an alternative multilingual reasoning enhancement method, which only leverages the \textbf{\textit{cross-lingual enhancement vectors}} in a reversed direction. 
We denote these vectors as the \textbf{\textit{latent english debiasing vectors}}, designed to mitigate English bias in latent space during understanding forward passing:
\begin{equation}
\boldsymbol{v}_{debias}^{(l)} = -\frac{1}{|\mathcal{X}|} \sum_{x \in \mathcal{X}} \Delta\boldsymbol{h}_x^{(l)}.
\label{eq:reasoning_vector_final}
\end{equation}
\begin{figure*}[!ht]
  \includegraphics[width=1.0\linewidth]{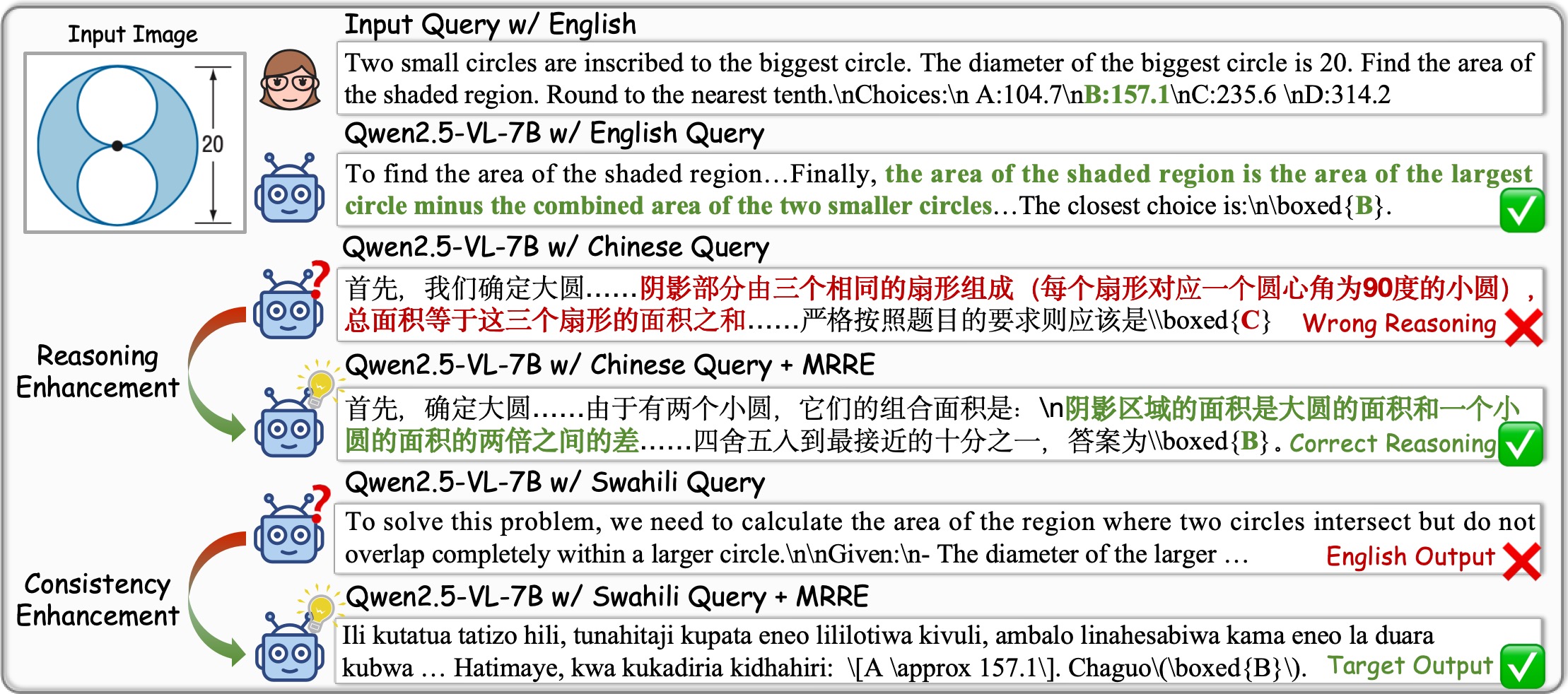}
  \caption{Case study of Qwen2.5-VL-7B-Instruct on the MathVerse benchmark.}
  \label{case0}
\end{figure*}

To validate this hypothesis, we conduct experiments with Qwen2.5-7B-Instruct and Qwen2.5-VL-7B-Instruct and apply \textbf{\textit{latent english debiasing vectors}} for each model. As illustrated in Table \ref{debiasing}, the MRRE \textit{Debias} method consistently improves cross-lingual performance on the Qwen2.5-7B-Instruct across all evaluated languages, and even surpasses the vanilla MRRE \textit{Vanilla} on Chinese, Thai, and Swahili. However, when applied to the Qwen2.5-VL-7B-Instruct, MRRE \textit{Debias} exhibits a consistent performance decline. According to prior study \cite{ye2025claim}, we posit that LVLMs rely predominantly on English data during the image–text alignment stage of pre-training, which endows English with stronger multimodal representations. 
Consequently, injecting \textbf{\textit{latent english debiasing vectors}} into non-English hidden states disrupts multimodal representations, leading to performance drops.

\begin{figure}
  \includegraphics[width=1.0\linewidth]{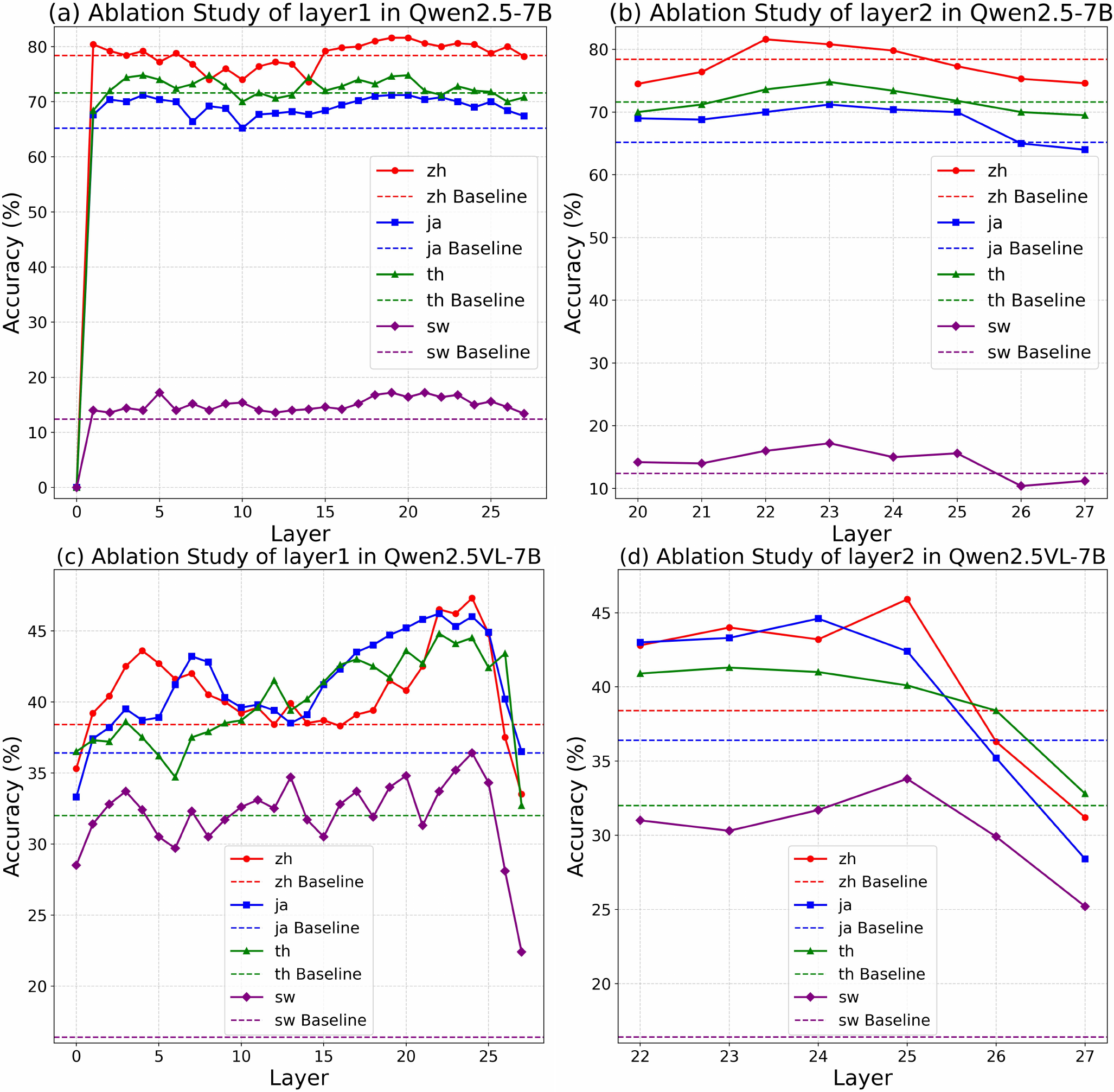}
  \caption{Analysis of layer selections of MRRE.}
  \label{ablation}
\end{figure}

\begin{table}
\centering
\small
\renewcommand{\arraystretch}{0.93}
\setlength{\tabcolsep}{0.7pt}

\begin{tabular}{l| S S S S S S S |S}
\toprule

\multicolumn{1}{c}{\textbf{Model}}  & {\textbf{Zh}} & {\textbf{Ja}} & {\textbf{Es}} & {\textbf{Ru}} & {\textbf{Fr}}  & {\textbf{Th}} & {\textbf{Sw}} &\textbf{LC} \\

\midrule
\rowcolor{gray!15} Qwen2.5-7B & 78.4 & 65.2 & 81.2 & 82.0 & 77.2  & 71.6 & 12.4 &84.7 \\
\quad + MRRE \textit{Vector1} &\cellcolor{green!15}\textbf{83.2}&\cellcolor{green!15}\textbf{73.7}&\cellcolor{green!15}\textbf{82.3}&\cellcolor{green!15}\textbf{85.2}&\cellcolor{green!15}\textbf{80.1}&\cellcolor{green!15}\textbf{75.2}&\cellcolor{green!15}\textbf{19.2}&26.2 \\
\quad + MRRE \textit{Vanilla} &81.6 &71.2 &81.2&84.7 &79.2 &74.8&17.2 &\cellcolor{green!15}\textbf{92.7} \\
\midrule
\rowcolor{gray!15} Qwen2.5VL-7B &58.1&52.0&62.2&57.9&59.6&51.1&16.7&94.1 \\
\quad + MRRE \textit{Vector1} &\cellcolor{green!15}\textbf{63.4}&\cellcolor{green!15}\textbf{62.3}&\cellcolor{green!15}\textbf{64.9}&\cellcolor{green!15}\textbf{64.2}&\cellcolor{green!15}\textbf{63.9}&\cellcolor{green!15}\textbf{62.5}&\cellcolor{green!15}\textbf{35.4}&30.0\\
\quad+ MRRE \textit{Vanilla}&62.3 &61.1 &64.2 &62.8 &61.2  & 57.3 &28.6&\cellcolor{green!15}\textbf{95.5}  \\

\bottomrule
\end{tabular}

\caption{Ablation studies of MRRE vectors.}
\label{vector_ablation}
\end{table}

\subsection{Analysis of Hyperparameters}
This subsection systematically examines the influence of the intervention layer1 $l$ and layer2 $l'$ for two designed vectors, respectively.
As shown in Figure \ref{ablation}, applying the first MRRE vector near layer 20 yields the best improvement in the model's reasoning ability. Given the first intervention at layer 20, applying the second vector near layer 23 achieves the next optimal boost. LVLMs exhibit a similar pattern, with the optimal layers being 22 and 24, respectively. As shown in Table~\ref{vector_ablation}, the second vector exhibits a trade-off strategy, which decreases little reasoning capabilities but significantly increases the consistency. Furthermore, we find MRRE achieves optimal results when $\alpha_1=1$ and $\alpha_2 = 0.75$. See Appendix \ref{moreablation} for more ablation studies of intervention strengths $\alpha_1$ and $\alpha_2$. 

\subsection{Case Study}
As shown in Figure \ref{case0}, 
MRRE not only enhances the model’s reasoning capability in non-English settings, but also improves the consistency between input and output languages. See Appendix \ref{morecase} for fine-grained case studies across LLMs and LVLMs.

\subsection{Efficiency Analysis}
To assess the practical deployability of MRRE, we analyze its computational overhead. 
\paragraph{Offline Preparation} The steering vectors are precomputed using a subset of 100 samples. This involves only element-wise subtraction and normalization, representing a negligible one-time cost. 
\paragraph{Inference Overhead} During inference, MRRE modifies the hidden states via two vector additions per token. We measure the latency using Qwen2.5-7B on a single NVIDIA A100 GPU. As reported in Table~\ref{tab:latency}, the impact on \textit{Time to First Token} (TTFT) and \textit{Time Per Output Token} (TPOT) is minimal ($<0.5$ms), confirming that MRRE maintains the high-speed generation characteristic.

\begin{table}[ht]
\centering
\caption{Inference latency comparison of the base model and MRRE.}
\label{tab:latency}
\begin{tabular}{lcc}
\toprule
\textbf{Method} & \textbf{TTFT (ms)} & \textbf{TPOT (ms)} \\ \midrule
Qwen2.5-7B      & 83.7               & 32.1               \\
+ MRRE          & 84.1 (\textcolor{red}{+0.4}) & 32.5 (\textcolor{red}{+0.4}) \\ \bottomrule
\end{tabular}
\end{table}


\section{Conclusion}
We propose a training-free inference-time method to enhance \textbf{M}ultilingual \textbf{R}easoning capabilities via \textbf{R}epresentation \textbf{E}ngineering (\textbf{MRRE}), which applies cross-lingual reasoning enhancement
vectors and target-language output anchoring vectors sequentially at specific layers of LLMs and LVLMs during forward passing. 
Comprehensive results demonstrate its effectiveness and generalizability.

\section*{Limitations}
One limitation of our work is that MRRE requires access to the internal representations of the model, making it infeasible for closed-source LLMs or LVLMs. Furthermore, due to the constraints of cost and resources, we only conduct experiments on these widely used benchmarks and models.

\section*{Acknowledgements}
Xiaocheng Feng is the corresponding author of this work.
We thank the anonymous reviewers for their insightful comments.
This work was supported by the National Natural Science Foundation of China (NSFC) (grant 62522603, 62276078), the Key R\&D Program of Heilongjiang via grant 2022ZX01A32, the Fundamental Research Funds for the Central Universities ( XNJKKGYDJ2024013 ) .


\bibliography{custom}

\clearpage

\appendix
\section{Translation Quality}
\label{trans}

To evaluate the translation quality of our constructed \textbf{MMathVerse} and \textbf{MMathVista} benchmarks, we sample 500 translated queries from each benchmarks for each language and back-translate them into English using Google Translate. The back-translated English queries are then input into LVLMs to test whether their predictions align with those generated from the original English queries. High prediction consistency indicates that the translated data maintains superior benchmark quality.
As shown in Table~\ref{trans_quality}, these results demonstrate the reliability of our constructed multilingual dataset. 
\begin{table}[!ht]
\setlength{\tabcolsep}{3.3pt}
\centering
\small
\begin{tabular}{l S S S S S S S S}
\toprule
Lang. & {\textbf{Zh}} & {\textbf{Ja}} & {\textbf{Es}} & {\textbf{Ru}} & {\textbf{Fr}} & {\textbf{De}} & {\textbf{Th}} & {\textbf{Sw}} \\
\midrule
APC&100.0&100.0&100.0&99.8&100.0&100.0&99.8&99.7 \\

\bottomrule
\end{tabular}
\caption{Average Predicted Consistency (APC, \%) MMathVerse and MMathVista across eight languages.}
\label{trans_quality}
\end{table}

\section{Experimental Details}
\label{exidetails}
In this section, we present the experimental details, including inference settings, design of experimental prompts, fine-grained results of MathVerse \& MathVista, and ablation of intervention strength.   
\subsection{Inference Settings}
To ensure that the reasoning parameters are better aligned with reasoning tasks and to guarantee the reproducibility of results, we carefully design inference settings for each model, as shown in Table~\ref{inference_setting}.
\begin{table}[!ht]
\setlength{\tabcolsep}{0.5pt}
\centering
\small
\begin{tabular}{l c}
\toprule
\textbf{Model} & \textbf{Setting} \\
\midrule
Qwen2.5-7B-Instruct& \makecell[c]{do\_sample=True,\\ temperature=0.7, \\ top\_p=0.8, top\_k=20}\\
\midrule
Qwen3-8B & \makecell[c]{do\_sample=True,\\ temperature=0.7, \\ top\_p=0.8, top\_k=20, \\enable\_thinking=False} \\
\midrule
Llama-3.1-8B- Instruct& do\_sample=True \\
\midrule
LLaVA-Onevision-7B & \makecell[c]{do\_sample=True,\\ temperature=0.6}\\
\midrule
Qwen2.5-VL-7B-Instruct &\makecell[c]{do\_sample=True,\\temperature=0.1, top\_p=0.001, \\repetition\_penalty=1.1}\\
\midrule
InternVL3.5-VL-8B-Chat &\makecell[c]{do\_sample=True,\\temperature=0.1, top\_p=0.001,\\ repetition\_penalty=1.1} \\
\bottomrule
\end{tabular}
\caption{Inference settings for each LLM and LVLM.}
\label{inference_setting}
\end{table}

\subsection{Design of Experimental Prompts}

In this section, we provide a detailed description of each type of prompt, along with intended purposes.

\paragraph{Language Forcing Prompts} are designed to enforce the model to generate outputs in the same language as the input. Following prior work \cite{wang2025polymath}, we adopt a similar strategy, with the detailed prompt contents provided in Table~\ref{language_prompts}.
\begin{table}[!ht]
\setlength{\tabcolsep}{0.01pt}
\centering
\small
\begin{tabular}{c c}
\toprule
\textbf{Lang.}  & \textbf{Prompts} \\
\midrule
En &Use English to think and answer.\\
\midrule
Zh &\begin{CJK}{UTF8}{gbsn}
使用中文进行思考和回答。
\end{CJK}
\\
\midrule
Ja & \begin{CJK}{UTF8}{gbsn}
日本語を使って考え、回答してください。
\end{CJK}\\
\midrule
Es &\begin{CJK}{UTF8}{gbsn}
Usa español para pensar y responder.
\end{CJK}\\
\midrule
Fr &\begin{CJK}{UTF8}{gbsn}
Utilisez le français pour penser et répondre.
\end{CJK}\\
\midrule
De &\begin{CJK}{UTF8}{gbsn}
Verwende Deutsch, um zu denken und zu antworten.
\end{CJK}\\
\midrule
Sw &\begin{CJK}{UTF8}{gbsn}
Tumia Kiswahili kufikiri na kujibu.
\end{CJK}\\
\bottomrule
\end{tabular}
\caption{Language forcing prompts contents.}
\label{language_prompts}
\end{table}

\paragraph{Prompts for MGSM and MSVAMP}
These two datasets primarily adopt free-form formats to evaluate the reasoning capabilities of LLMs. As shown in Table~\ref{llm_prompts},  we ask the model to first generate reasoning responses and then mark final answers using the \texttt{\textbackslash \texttt{\textbackslash boxed\{\}}} format.

\begin{table}[!ht]
\setlength{\tabcolsep}{4pt} 
\centering
\small
\begin{tabular}{c p{0.8\linewidth}} 
\toprule
\textbf{Lang.}  & \textbf{Prompts} \\
\midrule
En & Please first reason through the problem, then provide the final answer, expressed as a number using the \texttt{\textbackslash \texttt{\textbackslash boxed\{\}}} format. \\
\midrule
Zh & \begin{CJK}{UTF8}{gbsn}
请首先进行推理，然后给出最后的答案，用\texttt{\textbackslash \texttt{\textbackslash boxed\{\}}}的形式表示最后的数字。
\end{CJK} \\
\midrule
Ja & \begin{CJK}{UTF8}{gbsn}
まず推理を行ってください。その後、最終的な答えを\texttt{\textbackslash \texttt{\textbackslash boxed\{\}}}の形式で表示してください。
\end{CJK} \\
\midrule
Es & \begin{CJK}{UTF8}{gbsn}
Por favor, primero realice el razonamiento y luego dé la respuesta final, representando el número final en forma de \texttt{\textbackslash \texttt{\textbackslash boxed\{\}}}.
\end{CJK} \\
\midrule
Fr & \begin{CJK}{UTF8}{gbsn}
Veuillez d'abord raisonner, puis donner la réponse finale sous la forme \texttt{\textbackslash \texttt{\textbackslash boxed\{\}}}.
\end{CJK} \\
\midrule
De & \begin{CJK}{UTF8}{gbsn}
Bitte führen Sie zunächst Ihre Überlegungen durch und geben Sie dann die endgültige Antwort an. Verwenden Sie zur Darstellung der endgültigen Zahl die Form \texttt{\textbackslash \texttt{\textbackslash boxed\{\}}}.
\end{CJK} \\
\midrule
Sw & \begin{CJK}{UTF8}{gbsn}
Tafadhali fanya hoja kwanza, kisha toa jibu la mwisho kwa kuandika nambari ya mwisho katika umbo la \texttt{\textbackslash \texttt{\textbackslash boxed\{\}}}.
\end{CJK} \\
\bottomrule
\end{tabular}
\caption{Reasoning prompts for MGSM and MSVAMP.}
\label{llm_prompts}
\end{table}

\paragraph{Prompts for MathVerse and MathVista}
These two benchmarks are employed to evaluate the reasoning capability of LVLMs, covering two categories of tasks: (1) multiple-choice and (2) free-form. As shown in Table~\ref{lvlm_prompts}, we design tailored reasoning prompts to guide the model's response for multiple-choice queries. The prompts for free-form are the same as the prompts for MGSM and MSVAMP benchmarks.
\begin{table}[!ht]
\setlength{\tabcolsep}{2pt} 
\centering
\small
\begin{tabular}{c p{0.85\linewidth}} 
\toprule
\textbf{Lang.}  & \textbf{Prompts} \\
\midrule
En & Please first conduct reasoning, then answer the question and put the correct option letter into \texttt{\textbackslash \texttt{\textbackslash boxed\{\}}}, e.g., \texttt{\textbackslash \texttt{\textbackslash boxed\{A\}}}, \texttt{\textbackslash \texttt{\textbackslash boxed\{B\}}}, \texttt{\textbackslash \texttt{\textbackslash boxed\{C\}}}, \texttt{\textbackslash \texttt{\textbackslash boxed\{D\}}}, at the end.\\
\midrule
Zh & \begin{CJK}{UTF8}{gbsn}
请先进行推理, 然后回答问题, 并在最后将正确的选项字母填入\texttt{\textbackslash \texttt{\textbackslash boxed\{\}}}中, 例如：\texttt{\textbackslash \texttt{\textbackslash boxed\{A\}}}, \texttt{\textbackslash \texttt{\textbackslash boxed\{B\}}}, \texttt{\textbackslash \texttt{\textbackslash boxed\{C\}}}, \texttt{\textbackslash \texttt{\textbackslash boxed\{D\}}}. \end{CJK} \\
\midrule
Ja & \begin{CJK}{UTF8}{gbsn}
まず推論を行い、次に質問に答えて、正しいオプション文字を \texttt{\textbackslash \texttt{\textbackslash boxed\{\}}}に入れます。例: \texttt{\textbackslash \texttt{\textbackslash boxed\{A\}}},\texttt{\textbackslash \texttt{\textbackslash boxed\{B\}}},\texttt{\textbackslash \texttt{\textbackslash boxed\{C\}}},\texttt{\textbackslash \texttt{\textbackslash boxed\{D\}}}, 最後に。
\end{CJK} \\
\midrule
Es & \begin{CJK}{UTF8}{gbsn}
Primero realice el razonamiento, luego responda la pregunta y coloque la letra de la opción correcta en \texttt{\textbackslash \texttt{\textbackslash boxed\{\}}}, por ejemplo, \texttt{\textbackslash \texttt{\textbackslash boxed\{A\}}}, \texttt{\textbackslash \texttt{\textbackslash boxed\{B\}}}, \texttt{\textbackslash \texttt{\textbackslash boxed\{C\}}}, \texttt{\textbackslash \texttt{\textbackslash boxed\{D\}}}, al final.
\end{CJK} \\
\midrule
Fr & \begin{CJK}{UTF8}{gbsn}
Veuillez d'abord effectuer un raisonnement, puis répondre à la question et mettre la lettre d'option correcte dans \texttt{\textbackslash \texttt{\textbackslash boxed\{\}}}, par exemple, \texttt{\textbackslash \texttt{\textbackslash boxed\{A\}}}, \texttt{\textbackslash \texttt{\textbackslash boxed\{B\}}}, \texttt{\textbackslash \texttt{\textbackslash boxed\{C\}}}, \texttt{\textbackslash \texttt{\textbackslash boxed\{D\}}}, à la fin.
\end{CJK} \\
\midrule
De & \begin{CJK}{UTF8}{gbsn}
Bitte führen Sie zuerst eine Argumentation durch, beantworten Sie dann die Frage und setzen Sie am Ende den richtigen Optionsbuchstaben in \texttt{\textbackslash \texttt{\textbackslash boxed\{\}}}, z. B. \texttt{\textbackslash \texttt{\textbackslash boxed\{A\}}}, \texttt{\textbackslash \texttt{\textbackslash boxed\{B\}}}, \texttt{\textbackslash \texttt{\textbackslash boxed\{C\}}}, \texttt{\textbackslash \texttt{\textbackslash boxed\{D\}}}.
\end{CJK} \\
\midrule
Sw & \begin{CJK}{UTF8}{gbsn}
Tafadhali kwanza elekeza hoja, kisha ujibu swali na uweke herufi ya chaguo sahihi kwenye \texttt{\textbackslash \texttt{\textbackslash boxed\{\}}}, k.m., \texttt{\textbackslash \texttt{\textbackslash boxed\{A\}}}, \texttt{\textbackslash \texttt{\textbackslash boxed\{B\}}}, \texttt{\textbackslash \texttt{\textbackslash boxed\{C\}}}, \texttt{\textbackslash \texttt{\textbackslash boxed\{D\}}}, mwishoni.
\end{CJK} \\
\bottomrule
\end{tabular}
\caption{Multi-choice reasoning prompts for MathVerse and MathVista.}
\label{lvlm_prompts}
\end{table}


\subsection{Ablation of Intervention Strength}
\label{moreablation}
In this subsection, we conduct a detailed analysis of how the intervention strengths $\alpha_1$ and $\alpha_2$ influence the performance of MRRE. 

We first analyze the intervention strength of the cross-lingual enhancement vectors, $\alpha_1$, which aim to align non-English reasoning hidden states with English in the latent space. As shown in Table~\ref{a1}, the optimal performance is achieved when $\alpha_1 =1$, while weaker or stronger interventions fail to minimize the distribution gap in the latent space, limiting the effectiveness of MRRE.

Then we analyze the intervention strength of the cross-lingual enhancement vectors, $\alpha_2$, which aim to align English-like generating hidden states with the original non-English in the latent space. As shown in Table~\ref{a2}, the optimal performance is achieved when $\alpha_2 =0.75$, while weaker interventions may lead to English responses rather than target non-English responses and too strong interventions may result in performance drops in reasoning tasks. Notably, $\alpha _2$ functions as a trade-off strategy, effectively \textbf{balancing multilingual reasoning performance with input-output language consistency}. 
\begin{table}[!ht]
\setlength{\tabcolsep}{6pt}
\centering
\small
\begin{tabular}{l cccccc}
\toprule
$\alpha_1$ & 0.00 &0.25 & 0.50 & 0.75 & \textbf{1.00} & 1.25 \\
\midrule
Zh &78.4 &79.3&80.5 &82.3 &\textbf{83.2}&82.5\\
\midrule
Ja &65.2& 67.1& 68.3 &71.0 &\textbf{73.7}&70.8\\
\midrule
Th &71.6 &72.3 &73.1 &74.2 &\textbf{75.2} &73.9\\
\bottomrule
\end{tabular}
\caption{Ablation study of $\alpha_1$ on MGSM benchmark using Qwen2.5-7B-Instruct. Layer1 is set to be 20.}
\label{a1}
\end{table}

\begin{table}[!ht]
\setlength{\tabcolsep}{3pt}
\centering
\small
\begin{tabular}{l cccccc}
\toprule
$\alpha_2$ & 0.00 &0.50 & \textbf{0.75} & 1.00  \\
\midrule
Zh &83.2 (26.5) &82.5 (78.7) &\textbf{81.6 (99.2)} &79.5 (99.5)\\
\midrule
Ja &73.7 (19.8) &71.9 (69.9)&\textbf{71.2 (99.1)} & 70.0 (99.6)\\
\midrule
Th &75.2 (7.9)  &75.0 (54.1)&\textbf{74.8 (99.2)}& 72.1 (99.4) \\
\bottomrule
\end{tabular}
\caption{Ablation study of $\alpha_2$ on MGSM benchmark using Qwen2.5-7B-Instruct. The first numbers denote accuracies (\%) and the second numbers denote Language Consistency (LC, \%). Layer1 is set to be 20, $\alpha_1$ is set to be 1.0, and Layer2 is set to be 23.}
\label{a2}
\end{table}

\subsection{Fine-grained Results of MathVerse and MathVista Benchmarks}
To further elucidate how MRRE improves LVLM reasoning, we perform a fine-grained analysis across eight languages, five visual task families (\textit{text-dominant, text-lite, vision-integrated, vision-dominant, and vision-only}), and two domains (\textit{general} and \textit{math}). We disaggregate performance and consistency metrics by language, task category, and domain to identify where MRRE yields the largest gains; results are reported in Tables~\ref{fine_qwen}.

\section{Fine-grained Case Studies}
\label{morecase}
To better illustrate the improvements in reasoning capabilities and input-output language consistency achieved by MRRE, we provide additional comprehensive case studies.
\begin{table*}
   
\begin{tabular}{lccccccccc}
\toprule
\multicolumn{1}{c}{\multirow{2}{*}{\textbf{Model}} }& \multicolumn{6}{c}{\textbf{\textit{MMathVerse}}} & \multicolumn{3}{c}{\textbf{\textit{MathVista}}} \\
\cmidrule(lr){2-7} \cmidrule(lr){8-10}
 & {\textbf{T-D}} & {\textbf{T-L}} & {\textbf{V-I}} & {\textbf{V-D}} & {\textbf{V-O}} & {\textbf{Overall}} & {\textbf{General}} & {\textbf{Math}} & {\textbf{Overall}}  \\
 \midrule \multicolumn{10}{c}{Language-Zh}\\
\midrule
\rowcolor{gray!15}
Qwen2.5-VL-7B  &44.4&39.5&35.5&36.0&36.5&38.4&59.3&57.0&58.1\\
\quad + Language forcing &41.8&35.3&33.4&34.6&38.7&36.8&\cellcolor{green!15}\textbf{60.8}&59.8&60.0 \\
\quad + MRRE &\cellcolor{green!15}\textbf{54.2}&\cellcolor{green!15}\textbf{47.0}&\cellcolor{green!15}\textbf{43.8}&\cellcolor{green!15}\textbf{42.9}&\cellcolor{green!15}\textbf{41.9}&\cellcolor{green!15}\textbf{45.9}&60.7&\cellcolor{green!15}\textbf{63.7}&\cellcolor{green!15}\textbf{62.3}\\
\midrule
\multicolumn{10}{c}{Language-Ja}\\
\midrule\rowcolor{gray!15}
Qwen2.5-VL-7B & 42.8&35.3&33.5&34.5&35.0&36.4&55.0&49.4&52.0   \\
\quad + Language forcing &44.2&39.2&36.8&36.5&39.6&37.7&54.8&51.5&53.0     \\
\quad + MRRE &\cellcolor{green!15}\textbf{51.6}&\cellcolor{green!15}\textbf{46.2}&\cellcolor{green!15}\textbf{42.5}&\cellcolor{green!15}\textbf{42.4}&\cellcolor{green!15}\textbf{40.5}&\cellcolor{green!15}\textbf{44.6}&\cellcolor{green!15}\textbf{59.3}&\cellcolor{green!15}\textbf{62.6}&\cellcolor{green!15}\textbf{61.1}   \\
\midrule
\multicolumn{10}{c}{Language-Es}\\
\midrule\rowcolor{gray!15}
Qwen2.5-VL-7B & 49.5&43.5&40.9&38.1&40.4&42.5&63.0&61.5&62.2        \\
\quad + Language forcing &51.8&43.4&40.5&42.3&40.1&43.6&62.2&60.4&61.2   \\
\quad + MRRE &\cellcolor{green!15}\textbf{55.5}&\cellcolor{green!15}\textbf{49.5}&\cellcolor{green!15}\textbf{46.4}&\cellcolor{green!15}\textbf{45.1}&\cellcolor{green!15}\textbf{40.7}&\cellcolor{green!15}\textbf{47.5}&\cellcolor{green!15}\textbf{64.1}&\cellcolor{green!15}\textbf{64.3}&6\cellcolor{green!15}\textbf{4.2}               \\
\midrule
\multicolumn{10}{c}{Language-Ru}\\
\midrule\rowcolor{gray!15}
Qwen2.5-VL-7B & 45.1&39.0&35.8&35.8&37.6&38.6&60.4&55.7&57.9       \\
\quad + Language forcing &46.7&41.2&39.6&38.1&37.4&40.6&61.3&55.7&58.3   \\
\quad + MRRE  &\cellcolor{green!15}\textbf{52.0}&\cellcolor{green!15}\textbf{45.3}&\cellcolor{green!15}\textbf{41.8}&\cellcolor{green!15}\textbf{42.3}&\cellcolor{green!15}\textbf{42.4}&\cellcolor{green!15}\textbf{44.7}&\cellcolor{green!15}\textbf{62.8}&\cellcolor{green!15}\textbf{62.8}&\cellcolor{green!15}\textbf{62.8}              \\
\midrule
\multicolumn{10}{c}{Language-Fr}\\
\midrule\rowcolor{gray!15}
Qwen2.5-VL-7B & 43.0&38.6&35.8&36.8&36.4&38.1&61.5&58.0&59.6             \\
\quad + Language forcing&46.6&40.0&36.9&36.5&37.9&39.6&61.1&58.0&59.4    \\
\quad + MRRE  &\cellcolor{green!15}\textbf{51.9}&\cellcolor{green!15}\textbf{47.5}&\cellcolor{green!15}\textbf{44.7}&\cellcolor{green!15}\textbf{42.0}&\cellcolor{green!15}\textbf{42.9}&\cellcolor{green!15}\textbf{45.8}&\cellcolor{green!15}\textbf{62.2}&\cellcolor{green!15}\textbf{60.2}&\cellcolor{green!15}\textbf{61.2}              \\
\midrule
\multicolumn{10}{c}{Language-De}\\
\midrule\rowcolor{gray!15}
Qwen2.5-VL-7B & 42.5&38.2&32.5&33.4&36.0&36.5&58.9&57.6&58.2             \\
\quad + Language forcing &50.5&39.0&35.3&36.9&38.2&40.0&56.1&57.6&56.9   \\
\quad + MRRE  &\cellcolor{green!15}\textbf{52.9}&\cellcolor{green!15}\textbf{47.0}&\cellcolor{green!15}\textbf{42.9}&\cellcolor{green!15}\textbf{43.8}&\cellcolor{green!15}\textbf{41.6}&\cellcolor{green!15}\textbf{45.7}&\cellcolor{green!15}\textbf{63.5}&\cellcolor{green!15}\textbf{64.6}&\cellcolor{green!15}\textbf{64.1}              \\
\midrule
\multicolumn{10}{c}{Language-Th}\\
\midrule\rowcolor{gray!15}
Qwen2.5-VL-7B & 36.8&33.1&31.2&31,2&27.5&32.0&52.0&50.4&51.1             \\
\quad + Language forcing &50.3&43.8&41.8&40.6&41.0&43.5&55.9&51.9&53.7   \\
\quad + MRRE &\cellcolor{green!15}\textbf{56.7}&\cellcolor{green!15}\textbf{51.0}&\cellcolor{green!15}\textbf{43.8}&\cellcolor{green!15}\textbf{43.8}&\cellcolor{green!15}\textbf{46.9}&\cellcolor{green!15}\textbf{48.4}&\cellcolor{green!15}\textbf{58.7}&\cellcolor{green!15}\textbf{55.8}&\cellcolor{green!15}\textbf{57.3}              \\
\midrule
\multicolumn{10}{c}{Language-Sw}\\
\midrule\rowcolor{gray!15}
Qwen2.5-VL-7B & 15.1&13.8&11.8&14.0&27.3&16.4&17.6&15.9&16.7             \\
\quad + Language forcing &20.1&16.0&15.0&15.0&22.8&17.8&8.7&11.3&10.1   \\
\quad + MRRE &\cellcolor{green!15}\textbf{34.4}&\cellcolor{green!15}\textbf{32.5}&\cellcolor{green!15}\textbf{31.6}&\cellcolor{green!15}\textbf{32.4}&\cellcolor{green!15}\textbf{38.2}&\cellcolor{green!15}\textbf{33.8}&\cellcolor{green!15}\textbf{20.4}&\cellcolor{green!15}\textbf{35.6}&\cellcolor{green!15}\textbf{28.6}               \\
\bottomrule
\end{tabular}
\caption{Fine-grained Results on Qwen2.5-VL-7B-Instruct across eight languages.}
\label{fine_qwen}
\end{table*}

\clearpage

\begin{figure*}
  \includegraphics[width=1.0\linewidth]{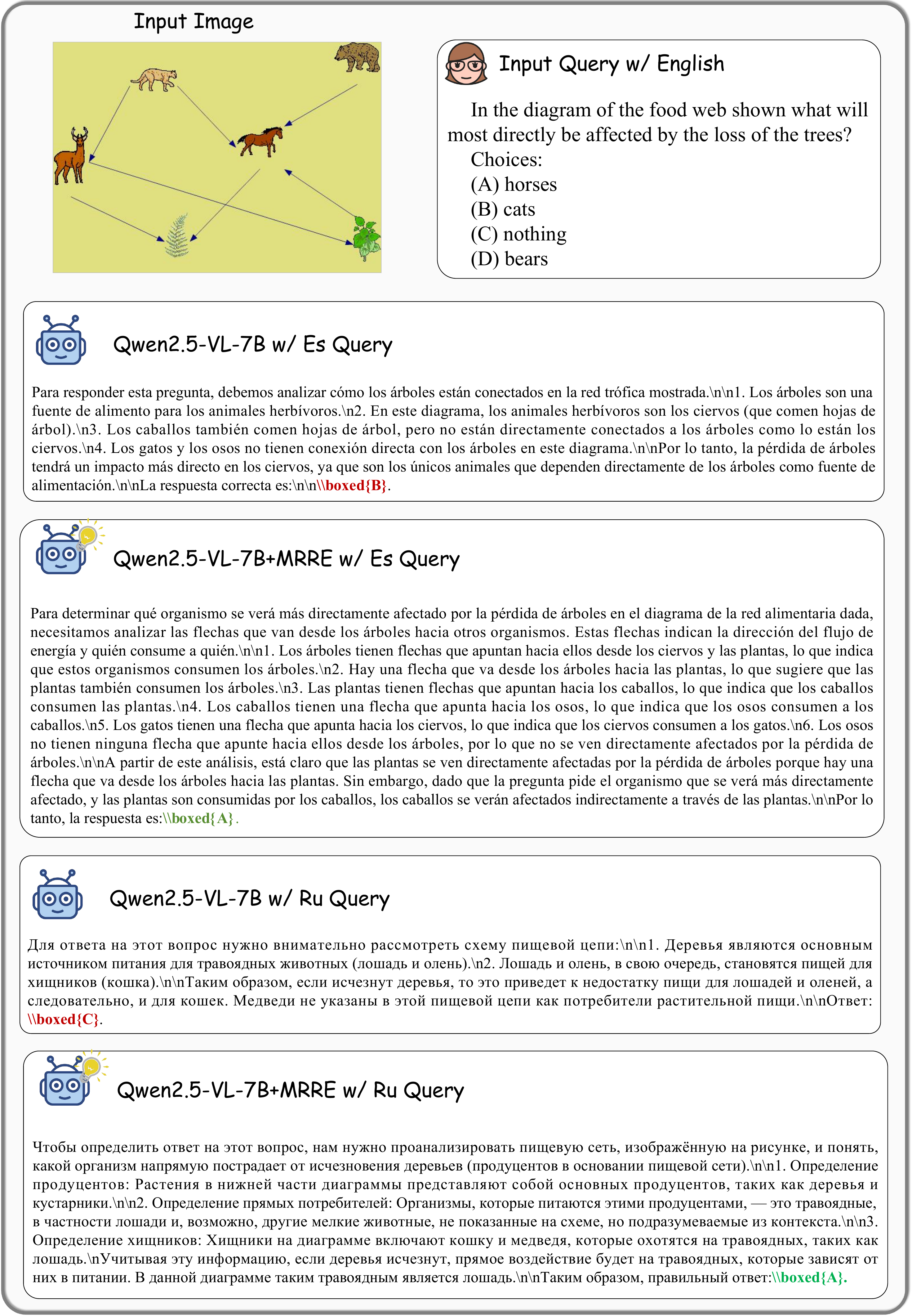}
  \caption{Case study of \textit{General} problem on Qwen2.5-VL-7B-Instruct. }
  \label{case_general}
\end{figure*}

\begin{figure*}
  \includegraphics[width=1.0\linewidth]{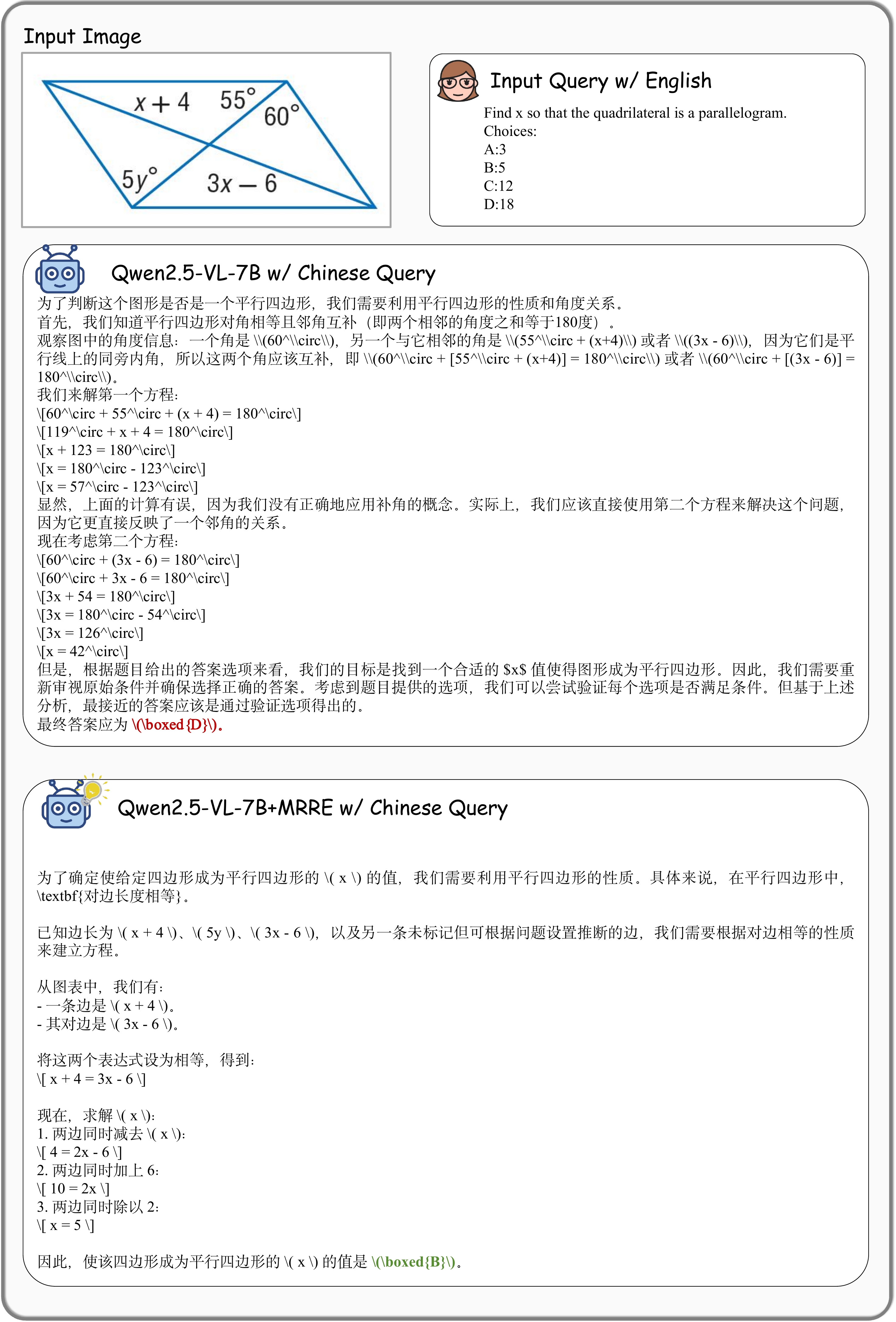}
  \caption{Case study of \textit{Text Lite} problem on Qwen2.5-VL-7B-Instruct. }
  \label{case1}
\end{figure*}

\begin{figure*}
  \includegraphics[width=1.0\linewidth]{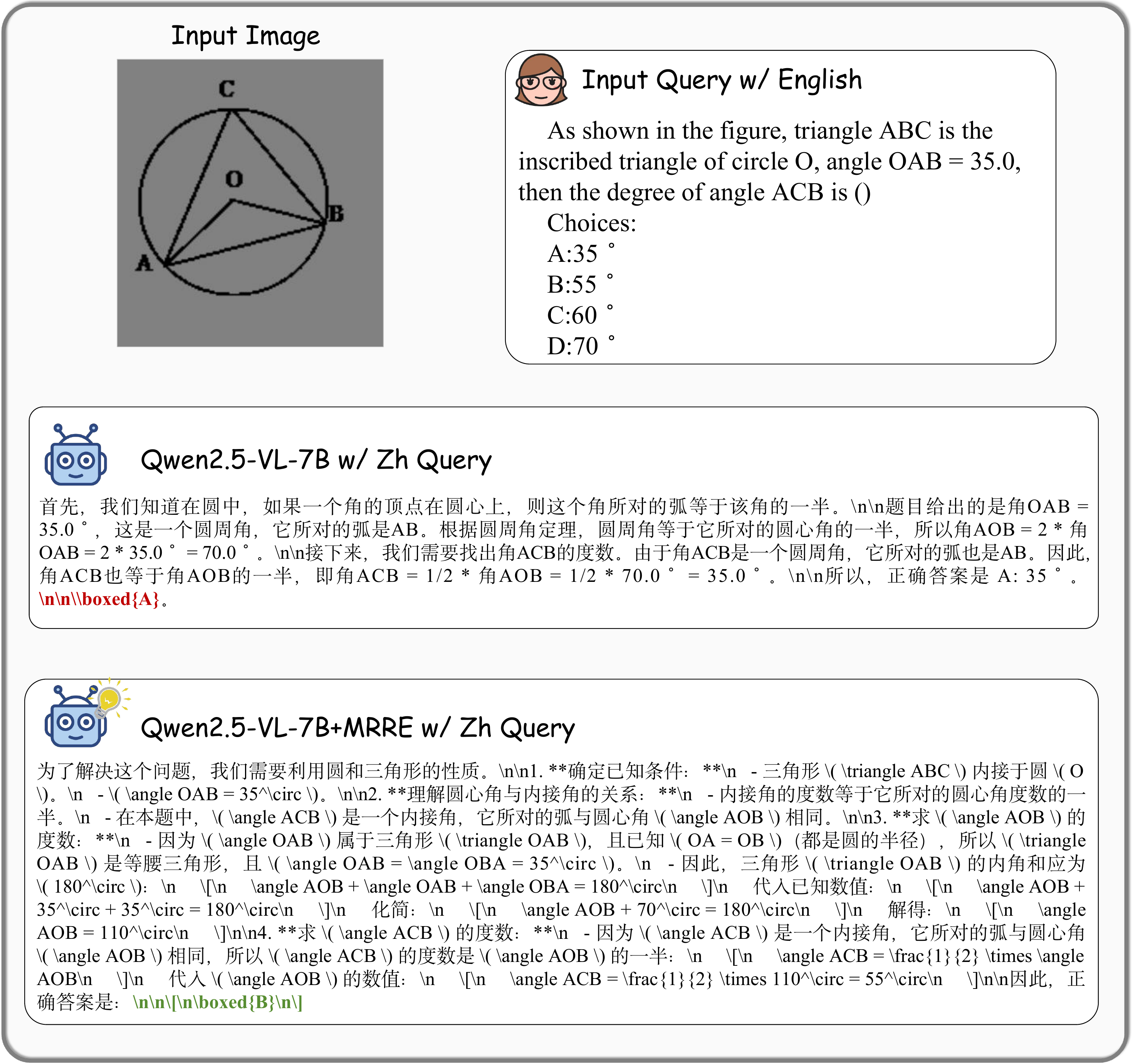}
  \caption{Case study of \textit{Text Dominant} problem on Qwen2.5-VL-7B-Instruct. }
  \label{case4}
\end{figure*}

\begin{figure*}
  \includegraphics[width=1.0\linewidth]{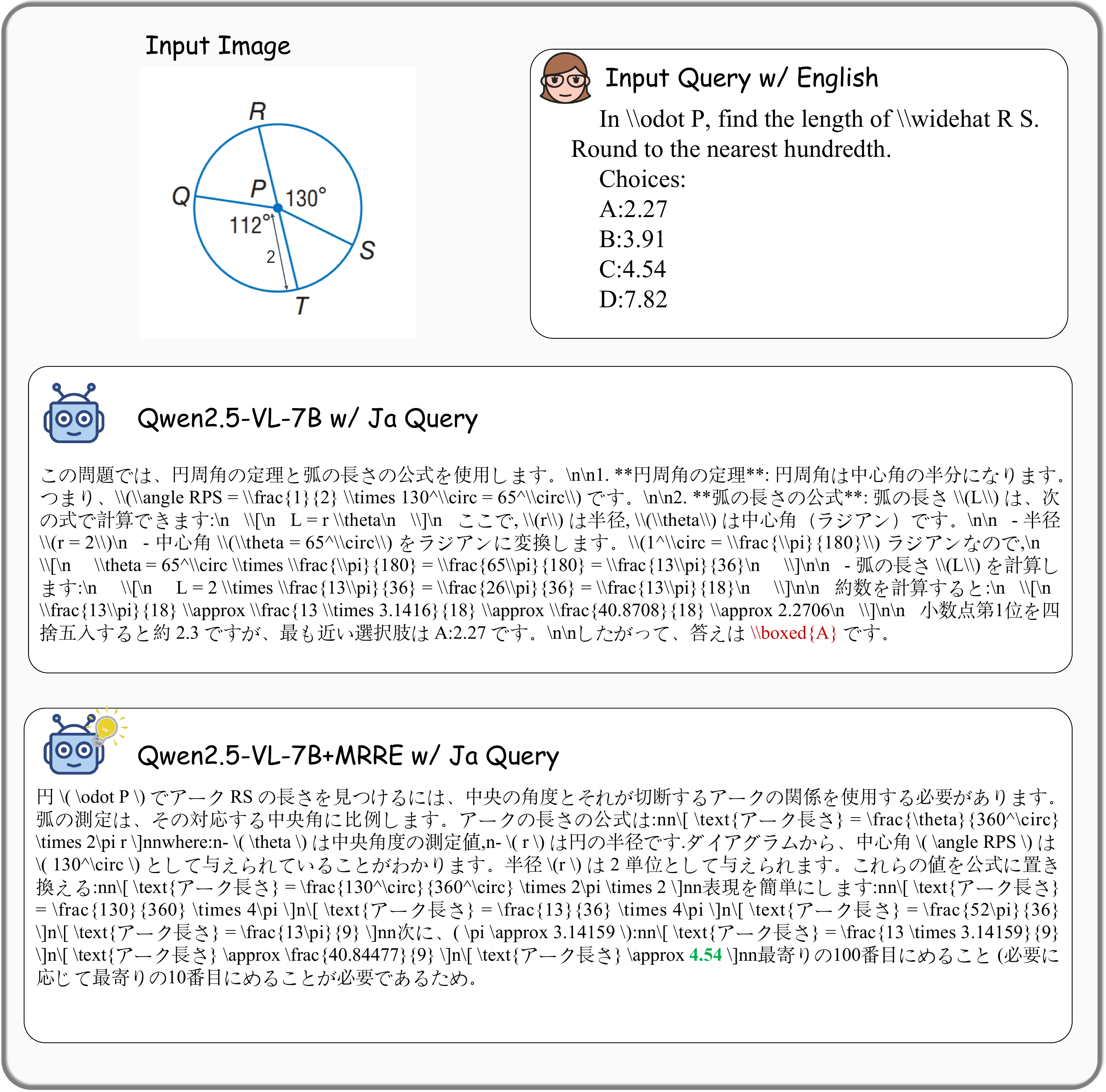}
  \caption{Case study of \textit{Vision Dominant} problem on Qwen2.5-VL-7B-Instruct. }
  \label{case2}
\end{figure*}

\begin{figure*}
  \includegraphics[width=1.0\linewidth]{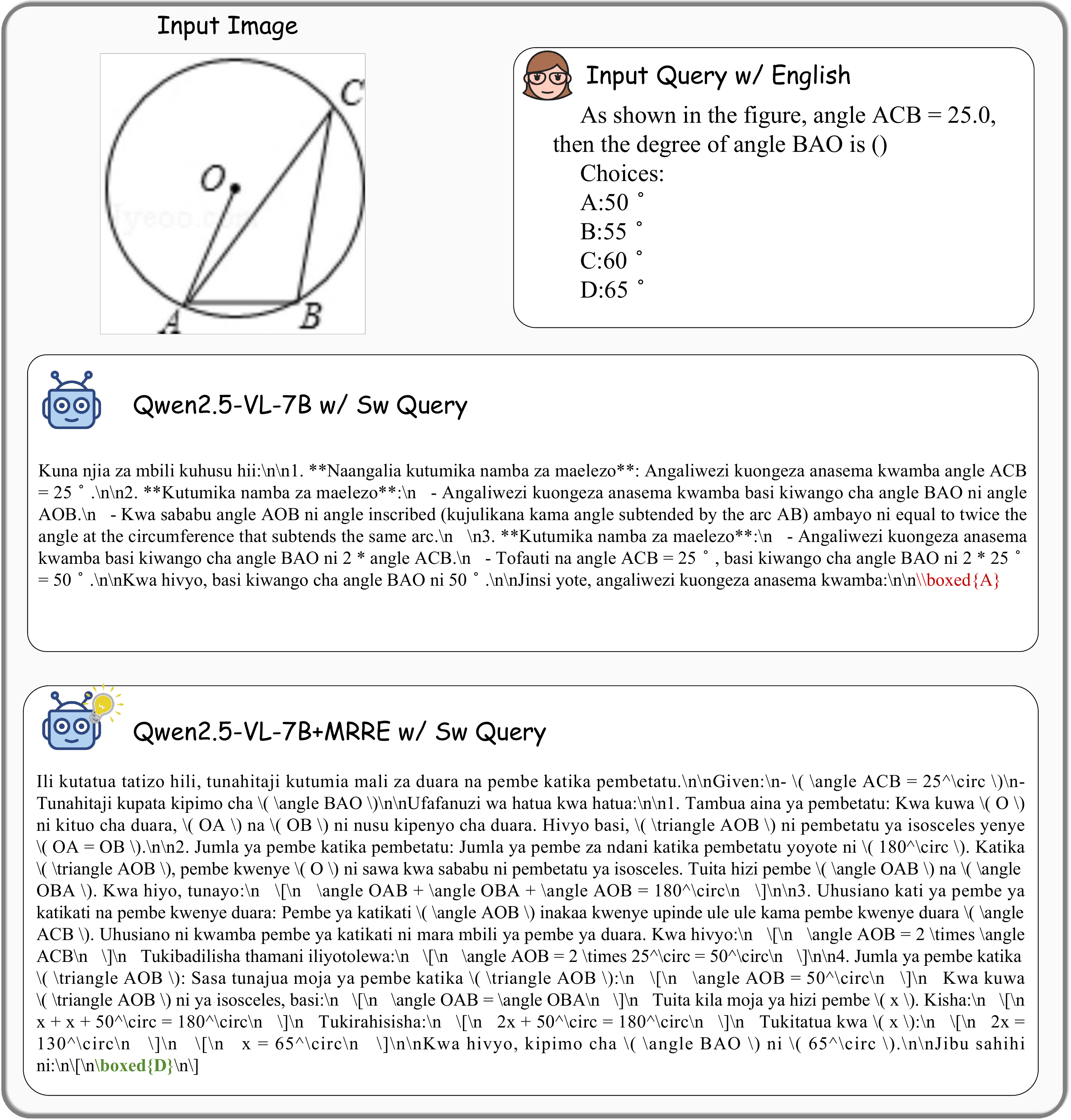}
  \caption{Case study of \textit{Vision Intensive} problem on Qwen2.5-VL-7B-Instruct. }
  \label{case3}
\end{figure*}

\begin{figure*}
  \includegraphics[width=1.0\linewidth]{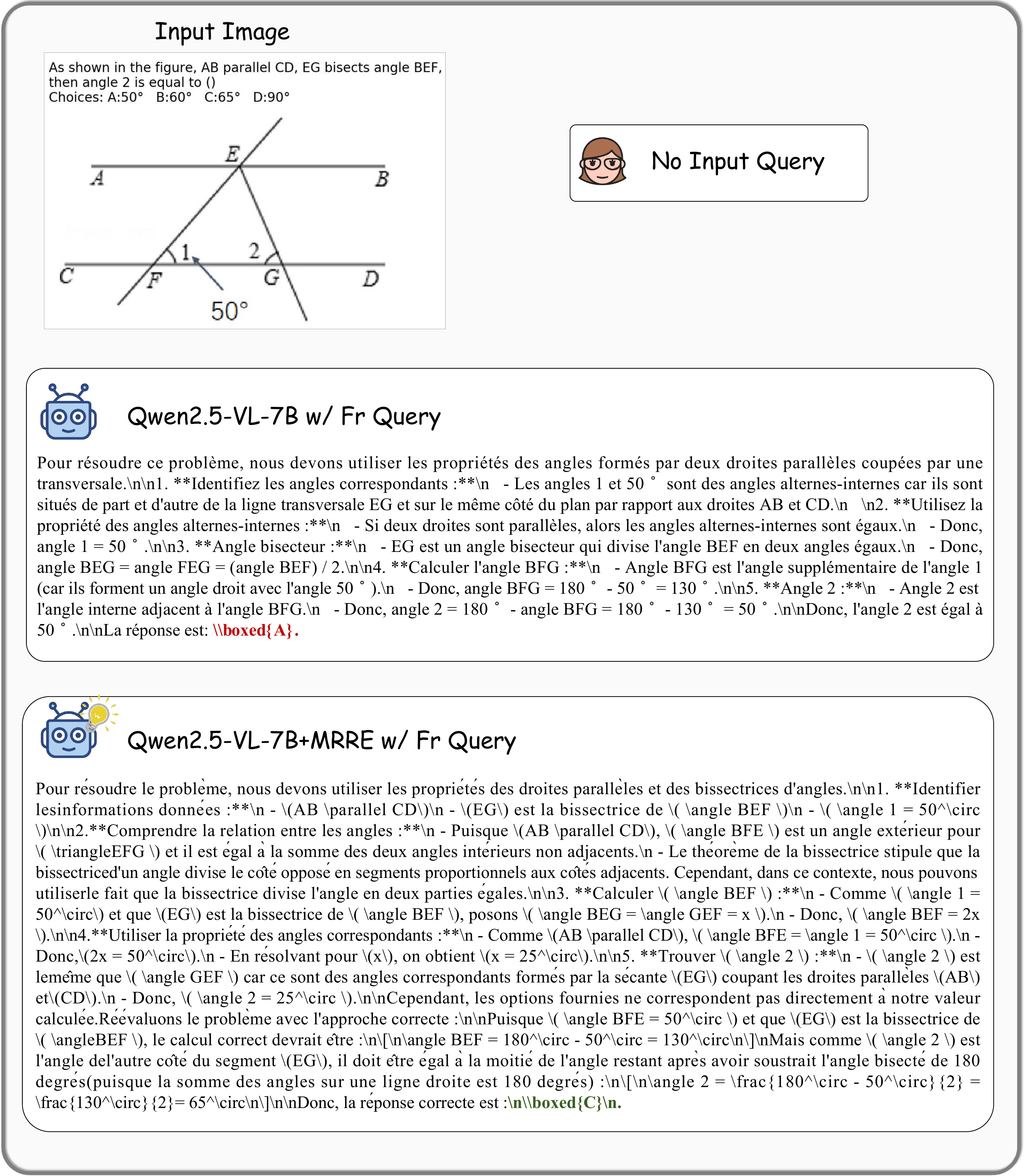}
  \caption{Case study of \textit{Vision Only} problem on Qwen2.5-VL-7B-Instruct. }
  \label{case5}
\end{figure*}

\clearpage

\section{Language Consistency (LC)}
To evaluate the model's proficiency in adhering to the specified target language, we define the \textit{Language Consistency} (LC) metric as the proportion of model responses successfully generated in the target language. Given the complexity of multilingual reasoning, where responses often contain a mixture of languages or technical symbols, we implement a character-weighted majority voting mechanism supported by FastText to ensure a noise-free evaluation.

As detailed in Algorithm~\ref{alg:lc}, our method first undergoes a preprocessing stage to strip LaTeX environments, mathematical commands, and URLs, preventing these elements from biasing the detector. A response is categorized as ``consistent'' only if the target language occupies more than $80\%$ of the total character count based on sentence-level decomposition.

\begin{algorithm}[ht]
\caption{Calculation of Language Consistency (LC)}
\label{alg:lc}
\SetKwInOut{Input}{Input}\SetKwInOut{Output}{Output}
\Input{Model responses $\mathcal{D}$ for target language $L$}
\Output{LC score (\%)}
\BlankLine
$total\_correct \leftarrow 0$\;
\ForEach{response $R \in \mathcal{D}$}{
    \tcp{Stage 1: Preprocessing}
    $R_{clean} \leftarrow$ Remove LaTeX math, commands, and URLs from $R$\;
    Clean extra symbols while preserving target scripts\;
    
    \tcp{Stage 2: Decomposition \& Detection}
    Split $R_{clean}$ into sentence-level chunks $\{C_1, C_2, \dots, C_n\}$\;
    $len_{valid} \leftarrow 0, len_{total} \leftarrow 0$\;
    \ForEach{chunk $C_i$}{
        Predict language $P_i$ and confidence $Prob_i$ using FastText\;
        \If{$P_i == L$ \textbf{and} $Prob_i \ge 0.8$}{
            $len_{valid} \leftarrow len_{valid} + \text{length}(C_i)$\;
        }
        $len_{total} \leftarrow len_{total} + \text{length}(C_i)$\;
    }
    
    \tcp{Stage 3: Sample Decision}
    $Ratio \leftarrow len_{valid} / len_{total}$\;
    \If{$Ratio > 0.8$}{
        $total\_correct \leftarrow total\_correct + 1$\;
    }
}
\Return{$(total\_correct / |\mathcal{D}|) \times 100$\%}\;
\end{algorithm}

\section{Usage of LLMs}
We used GPT-4o \cite{hurst2024gpt} to assist in language refinement and readability improvement of the manuscript. All ideas, experiments, analyses, and conclusions are developed and verified by the authors.

\end{document}